\documentclass[journal,compsoc]{IEEEtran}
\usepackage{amsmath,amsfonts}
\usepackage{array}
\usepackage[caption=false,font=normalsize,labelfont=sf,textfont=sf]{subfig}
\usepackage{textcomp}
\usepackage{stfloats}
\usepackage{url}
\usepackage{verbatim}
\usepackage{graphicx}
\def\BibTeX{{\rm B\kern-.05em{\sc i\kern-.025em b}\kern-.08em
    T\kern-.1667em\lower.7ex\hbox{E}\kern-.125emX}}
\usepackage{balance}

\usepackage{times}
\usepackage{soul}
\usepackage{url}
\usepackage{graphicx}
\usepackage{amsmath}
\usepackage{float}
\usepackage{amsthm}
\usepackage{multirow}
\usepackage{makecell}
\usepackage[T1]{fontenc}
\usepackage{amsfonts}
\usepackage{amssymb}
\usepackage{algpseudocode}
\usepackage{graphicx}
\usepackage{amsthm}
\usepackage{float}
\usepackage{multirow}
\usepackage{color}
\usepackage{stfloats}
\usepackage{epstopdf}
\usepackage{array}
\usepackage{verbatim}
\usepackage{booktabs}
\usepackage{graphicx}

\usepackage{fdsymbol}
\usepackage[ruled,linesnumbered]{algorithm2e}

\usepackage[table,xcdraw]{xcolor}
\usepackage[bookmarks=true, colorlinks, linkcolor=blue, urlcolor=blue, citecolor=blue]{hyperref}

\begin{document}

\title{Mitigating Prior Errors in Causal Structure Learning: A Resilient Approach via Bayesian Networks}
\author{Lyuzhou Chen, Taiyu Ban, Xiangyu Wang$^*$, Derui Lyu, Huanhuan Chen$^*$

\thanks{This research was supported in part by the National Key R\&D Program of China (No. 2021ZD0111700), in part by the National Nature Science Foundation of China (No. 62137002, 62406302, 62176245), in part by the  Natural Science Foundation of Anhui province (No. 2408085QF195), in part by the Fundamental Research Funds for the Central Universities under Grant WK2150110035.}
\thanks{Lyuzhou Chen, Taiyu Ban, Xiangyu Wang, Derui Lyu and Huanhuan Chen are with the School of Computer Science and Technology, University of Science and Technology of China, 96 Jinzhai Rd, Hefei 230026, China.}
\thanks{$^*$\textit{Corresponding authors}: Xiangyu Wang (sa312@ustc.edu.cn) and Huanhuan Chen (hchen@ustc.edu.cn).}
}

\markboth{IEEE TRANSACTIONS ON PATTERN ANALYSIS AND MACHINE INTELLIGENCE}%
{How to Use the IEEEtran \LaTeX \ Templates}

\IEEEtitleabstractindextext{
\begin{abstract}
Causal structure learning (CSL), a prominent technique for encoding cause-and-effect relationships among variables, through Bayesian Networks (BNs). Although recovering causal structure solely from data is a challenge, the integration of prior knowledge, revealing partial structural truth, can markedly enhance learning quality. However, current methods based on prior knowledge exhibit limited resilience to errors in the prior, with hard constraint methods disregarding priors entirely, and soft constraints accepting priors based on a predetermined confidence level, which may require expert intervention. To address this issue, we propose a strategy resilient to edge-level prior errors for CSL, thereby minimizing human intervention. We classify prior errors into different types and provide their theoretical impact on the Structural Hamming Distance (SHD) under the presumption of sufficient data. Intriguingly, we discover and prove that the strong hazard of prior errors is associated with a unique acyclic closed structure, defined as ``quasi-circle''. Leveraging this insight, a post-hoc strategy is employed to identify the prior errors by its impact on the increment of ``quasi-circles''. Through empirical evaluation on both real and synthetic datasets, we demonstrate our strategy's robustness against prior errors. Specifically, we highlight its substantial ability to resist order-reversed errors while maintaining the majority of correct prior.
\end{abstract}

}

\maketitle
\IEEEdisplaynontitleabstractindextext
\IEEEpeerreviewmaketitle

\section{Introduction}
\label{section_introduction}

\IEEEPARstart{C}{ausal} structure learning (CSL), the task to explore ``why'', forms a critical focus in the novel generation of pattern analysis and machine intelligence \cite{zhang2021survey,xin_wang,yu2019multi,chu2023machine,9756301}. The purpose of CSL is to obtain a Bayesian Network (BN) (or a Directed Acyclic Graph (DAG) if only structural dependencies are focused on), analyzing cause-and-effect relationships among variables \cite{scanagatta2019survey,jiang2019joint,li2014sparse}. Since the observed data may contain noise and bias, causal structure inferred from data alone may be imprecise. Therefore, some works propose employing prior constraints to reveal partial structural truth, rectify incorrect causal assertions, and guide the learning process\cite{de2011efficient,wang2023accurate,9815029}, which motivates the exploration of priors to solve causality problems.

However, ensuring the correctness of priors presents a new challenge. Indeed, prior errors can severely impact the accuracy of the inferred causal graph, not only affecting the relationships between the involved variables but also spreading to the surrounding graph structure of involved variables, potentially leading to incorrect inferences of multiple causal relationships.

Alarmingly, this risk is often overlooked by existing CSL methods. Current prior-based CSL methods fall into two categories: hard constraint and soft constraint approaches \cite{constantinou2021impact,de2008improving}, where prior-based hard constraint methods assume the correctness of the input priors \cite{asvatourian2020integrating} and does not consider this issue. Even soft constraint methods, which balance between data distribution and priors, essentially choose between data-based and prior-based results based on the confidence of the priors which is assumed correct \cite{9583890,9018179,jiang2017scalable}. When transitioning to real-world conditions for obtaining priors, the correctness of priors or their confidence is easily lost, 
especially when using non-expert-driven automated systems to obtain priors from knowledge sources such as knowledge graphs (KG) or large language models (LLMs) \cite{long2023can,ban2023causal,kiciman2023causal,ban2023query}. We argue for a skepticism towards priors and strive for a fundamental correction of priors, guiding causal structure learning with accurate priors instead of introducing incorrect information into the causal learning process.

To tackle this challenge, this paper aims to design a prior correction strategy resistant to unreliable priors. Specifically, we first investigate the mechanism by which prior errors affect the learned structure and demonstrate, both theoretically and experimentally, the specific impact of prior errors on the resulting causal graph. Notably, we highlight that priors that reverses the order of variables are particularly damaging. Fortunately, we show that this damage to the causal graph induces to generate a unique acyclical closed structure, termed a quasi-circle. This quasi-circle structure indicate the presence of prior errors. Consequently, we design a post-hoc checking procedure that utilizes quasi-circles to reveal prior errors. Excitingly, this strategy and the specific CSL method is separate and can be seamlessly integrated into existing prior-based CSL methods.

This paper applies the proposed prior correction strategy to classic and trendy methods in the field of CSL, respectively. Extensive empirical evaluations are conducted towards these method on both synthetic and real-world causal structure. 
Our experimental results indicate that, compared with the basic method, the improved method generally has better performance under different prior situations and various experiment parameters. Comparison with different methods also illustrates the superiority of the strategy proposed in this article in resisting prior errors.

The remainder of this paper is organized as follows. Section \ref{section_motivation} reviews various existing works that use priors to assist structure learning. Section \ref{section_preliminary} formally introduces the concepts and terminology covered in this paper, as well as the classification of prior errors. Section \ref{section_method} theoretically illustrates the influence of priors on learned structure and the connection between order-reversed priors and quasi-circles. Section \ref{section_algorithm} presents the proposed post-hoc strategy. Section \ref{section_experiment} evaluates proposed strategy experimentally. Section \ref{section_conclusions} draws conclusions.

\section{Motivation}
\label{section_motivation}

\subsection{Principal Causal Structure Learning Methods}
\label{section_Principal_CSL_Methods}

From a research perspective, causal structure learning encompasses three main approaches: score-based methods, constraint-based methods, and gradient-based methods \cite{vowels2022d}. Score-based methods, represented by Hill-Climbing (HC) \cite{kitson2023survey} and CaMML \cite{o2006causal}, utilize scoring functions to evaluate the fit of different DAGs with observed data. As one of the most classic CSL methods, the HC method performs greedy search in DAG space. It initiates with an empty graph and iteratively modifies the current graph—through edge additions, deletions, or reversals—to greedily enhance the congruence between the graph structure and the observed data. This process continues until any further modification would result in a decreased match.

Constraint-based methods, exemplified by the PC algorithm \cite{tsagris2019bayesian}, search for conditional independencies among variables within the data to infer causal structure. Gradient-based methods, with NOTEARS \cite{zheng2018dags} as a representative, leverage the optimization process to directly find the optimal solution in the space of edge weight parameters of DAGs. In addition to these, there are hybrid approaches, such as the widely used MMHC (Max-Min Hill-Climbing) \cite{tsamardinos2006max}, which combines elements of both score-based and constraint-based methods.

\subsection{Combination of Priors and Causal Learning}
Depending on how priors are used, causal structure learning methods bifurcates into soft constraints and hard constraints.  
Soft constraints not only evaluate the fit of different DAGs with observed data, but also evaluate the consistency between the DAGs and the prior \cite{castelo2000priors,amirkhani2016exploiting}.
For instance, Borboudakis \textit{et al.} \cite{borboudakis2014scoring} integrate priors on the presence or absence of specific paths in real networks into the scoring function. Eggeling \textit{et al.} \cite{eggeling2019structure} propose a global structure prior, setting the prior score according to the maximum number of parents of candidate graph nodes, fostering sparsity in the learned graph. Fundamentally, these methods aim to adjust the score of certain graphs in the search space to facilitate the discovery of graphs satisfying given priors.

Hard constraints, on the other hand, compel candidate graphs to adhere to specific properties or structure dictated by the priors \cite{borboudakis2012incorporating}. The most straightforward application of hard constraints is to dismiss candidate graphs that fail to satisfy the prior during the search process \cite{de2007bayesian,de2009structure}. To achieve a more accurate search, some works traverse the possible graph globally. For instance, Chen \textit{et al.} \cite{chen2016learning} suggest a tree structure that meets hard constraints by reducing branches. Works such as \cite{wang2021learning,li2018bayesian} share a similar idea. Some methods ascertain whether hard constraints are required during the learning process. For example, the algorithm proposed by Cano \textit{et al.} \cite{cano2011method} solicits expert opinion on whether an edge with probability close to 0.5 truly exists. Compared to soft constraints, hard constraints offer greater interpretability and ease of implementation.

\subsection{Challenges and Innovations in Addressing Prior Errors in Causal Structure Learning}
Despite the numerous studies investigating prior-based methods, a vast majority of these methods invariably presume the correctness of priors or the existence of prior confidence, neglecting the identification of prior errors as a potential problem. It is, in fact, sensible to moderate assumptions regarding the quality of the prior. In real environments, the priors provided by experts or derived from knowledge sources might contain errors or conflicts, posing a significant hurdle for methods that perform optimally only when the priors are entirely accurate. Specifically, for hard constraints, which necessitate that the result conforms to the prior, prior error will foreclose the opportunity of achieving a wholly accurate graph. Additionally, the influence of prior error may extend beyond the node pairs related to the prior error, potentially affecting the local or even global network structure. Soft constraints naturally have the capacity to exclude prior errors to an extent, as the highest score is not necessarily correlated with satisfying the prior. However, soft constraints do not identify prior errors but instead seek an optimal solution that fulfills more high-scoring graph structure. Consequently, this capacity may rely on a large proportion of correct priors. If all priors are erroneous, soft-constraint methods may still select a subset of prior errors, particularly if those priors are logically consistent.

A dearth of prior-based methods explores the identification of prior errors. Aside from soft-constraint methods that may discard part of the priors, the work of Amirkhani \textit{et al.} \cite{amirkhani2016exploiting} is one of the rare studies that explicitly considers prior errors. The authors model soft-constrained prior beliefs and integrate them into the score calculation. Through multiple iterations and optimization, they derive prior confidence and the optimal result of the Bayesian network. While for hard constraints, to the best of our knowledge, this paper pioneers the exploration of identifying prior errors. Herein, prior errors are segregated into different types based on their characteristics, and the impact of various prior errors on the result graph is probed. Subsequently, this paper introduces a unique acyclic closed structure, namely quasi-circles, to perform a post-hoc detection on order-reversed prior errors, which exhibit the most substantial influence on the SHD.

\section{Preliminary}
\label{section_preliminary}

This section first introduces the formal definitions of Bayesian networks and causal structure learning. Subsequently, we introduce several concepts, terminologies and properties that will facilitate the ensuing discussion. Lastly, we introduce a classification for prior errors, which is first defined in this paper.

\subsection{Background}

\textbf{Bayesian Network.} A Bayesian Network (BN) $B$ is a triple $(G,\mathbf{X},p)$, where $G=(\mathbf{V},\mathbf{E})$ is a directed acyclic graph (DAG) comprising a set of nodes $\mathbf{V}$ and a set of edges $\mathbf{E}\subset \mathbf{V}\times \mathbf{V}$ \cite{7364252,9079582}. $\mathbf{X}=\{X_1,X_2,...,X_n\}$ denotes a set of random variables, each corresponding to a node in $V$. In this paper, nodes in $V$ are also represented using the symbols of the corresponding random variables in $\mathbf{X}$. $p$ is the joint distribution of random variables, typically required to satisfy the faithfulness assumption, which postulates that $p$ contains only the independence conditions in $G$.

\textbf{Causal Structure Learning (CSL).} Suppose $P=\{e_1,e_2,...,e_n\}$ denotes the set of priors, where each prior $e$ is expressed in the form of $(X_i, X_j)$, that is, the prior can be viewed as a directed edge in DAG $G$. The prior $e=(X_i, X_j)$ is considered correct if edge $(X_i, X_j)$ exists in the DAG $G$; otherwise, it is erroneous. Let $D$ denote the dataset, where each sample is independently drawn from the joint distribution $p$. The goal of CSL is to discover a DAG that best fits the dataset $D$, thereby uncovering the structure of the BN. This paper goes a step further, aiming to identify and amend prior errors in $P$ and find the Bayesian network that best fits both the dataset $D$ and the modified prior set $P$.

\subsection{Terminology}
To facilitate more in-depth discussions, we clarify certain terminologies related to graph theory and CSL.

\begin{figure}[htbp]
\centering
\includegraphics[width=0.49\textwidth]{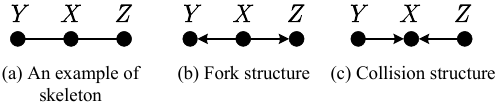}
\caption{Graph structure addressed in this paper.}
\label{terminology}
\end{figure}

In this paper, a \textit{path} refers to a graph structure composed of connected directed edges sharing the same direction, whose length equals the number of edges. The \textit{skeleton} corresponding to a path denotes the structure of the path after removing the direction, as demonstrated in Fig. \ref{terminology}(a).  Fig. \ref{terminology}(b) presents a \textit{fork} and Fig. \ref{terminology}(c) depicts a \textit{collision}. Note that in the collision structure, $Y$ and $Z$ are not independent given $X$. This property will be utilized in the following discussions.

This paper uses following expressions to indicate that graph $G$ contains a specific structure for convenience:
$$Struc \in G$$
where $Struc$ could be an edge, a path, a quasi-circle defined below, etc. In addition, for a node $X$ in graph $G$, $Pa^G(X)$ is used to denote the set of parent nodes of $X$ in graph $G$.

Score-based methods employ scoring functions to evaluate the reasonableness of the learned graph. The utilized scoring function is denoted by $S(\cdot,\cdot)$. Commonly used scoring functions include BIC (Bayesian Information Criterion) \cite{neath2012bayesian}, BDeu (Bayesian Dirichlet equivalent uniform) \cite{suzuki2017theoretical}, MDL (Minimum Description Length) \cite{hansen2001model}, etc.. These scoring functions possess a property called local consistency \cite{malinsky2019learning,gao2018parallel}:

\textbf{Definition (Local Consistency):}
Given a dataset $D$ consisting of sufficient records that are i.i.d (Independent Identically Distributed) samples from some distribution $p$. Let $G$ be any DAG, and let $G'$ be the DAG that results from adding the edge $X_j\longrightarrow X_i$. $Pa_i^G$ is the parents of $X_i$ in $G$. A scoring criterion $S(G, D)$ is locally consistent if the following two properties hold:

1. If $X_j \nVbar X_i | Pa_i^G$ in $p$, then $S(G' , D) > S(G, D)$

2. If $X_j \Vbar X_i | Pa_i^G$  in $p$, then $S(G' , D) < S(G, D)$

Local consistency indicates that if there is a direct causality between two nodes $X_i$, $X_j$, a scoring function with local consistency will favor adding an edge between these two nodes. Conversely, if $X_i$ and $X_j$ are independent given the currently determined set of parent nodes of node $X_i$, the scoring function will not favor adding $X_j$ as the new parent of $X_i$. This property proves highly valuable for the discussion and proof in the Section \ref{section_method}.

\subsection{Classification of Priors}

The effect of different prior errors on CSL may differ, which is related to the relationship of nodes in actual Bayesian networks. Accordingly, this paper introduces a novel division and definition of four types of prior errors:

\textbf{Definition (Classification of Priors):}
Prior errors can be classified into four categories:

\begin{itemize}
    \item \textbf{Reversed Direct Prior:}
    Prior that treats a \textbf{parent} of the node as it's children
    \item \textbf{Reversed Indirect Prior:}
    Prior that treats an \textbf{ancestor} of the node as it's children
    \item \textbf{Indirect Prior:}	
    Prior that treats a \textbf{grandchild} of the node as it's children
    \item \textbf{Irrelevant Prior:}	
    Prior that treats the \textbf{other node} as it's children
\end{itemize}

Any prior can, and can only, be categorized into one of the four types above. Both reversed indirect priors and reversed direct priors can lead to CSL results that significantly deviate from the true DAG $G$. This is primarily because these two types of priors invert the causal order. In fact, during causal learning, a prior that reverses the causal order changes the parent or child of some nodes, potentially disrupting the independence condition between nodes. Another possible scenario is that under the constraint that DAGs cannot contain circles, priors that reverse the causal order may further lead to misorientation of adjacent edges. From this viewpoint, the reversed direct prior and the reversed indirect prior can be collectively referred to as order-reversed prior errors, while the indirect prior can be referred to as order-consistent prior error. The remaining two types of priors, irrelevant priors and indirect priors, generally have little impact on the learning results. Therefore, it is crucial to identify reversed indirect priors and reversed direct priors and mitigate their influence on CSL. 
For the conclusions about these four types of prior errors, we will provide theoretical proof, and also confirm them through experimental verification.

\section{Method}
\label{section_method}

This section first defines a special type of graph structure, called quasi-circle, or QC for short. Using it as a tool, we explore the impact of each type of prior errors on causal structure learning and provide theoretical justification for the conclusions drawn. These findings will subsequently guide quasi-circle-based prior correction strategy, as described in the Section \ref{section_algorithm}.

\subsection{Quasi-circle}
\label{section_quasi_circle}

A quasi-circle is a causal graph pattern in which an ancestor node influences a descendant node through two independent causal paths. For example, consider this real-world scenario: educational attainment affects employment status, which in turn affects physical health; simultaneously, educational attainment affects health literacy, which in turn affects physical health. At this time, education level, employment status, health literacy and physical health form a quasi-circle. More precisely, we have the following definition:

\textbf{Definition (Quasi-circle):} Let $l_1$ and $l_2$ be two paths. If the head and tail nodes of $l_1$ and $l_2$ are identical, while the remaining nodes differ, the structure formed by $l_1$ and $l_2$ is termed a quasi-circle.

The quasi-circle and the circle are identical in terms of skeleton. A quasi-circle can be thought of as the result of flipping the direction of a continuous part of the circle. Since a DAG cannot contain circles, a quasi-circle can also be considered the closest structure to a circle that can appear in a DAG. Locally, the two paths in the quasi-circle form a fork at the head and a collision at the tail, so the quasi-circle must contain a fork, a collision, and at least one chain.

\begin{figure}[htbp]
    \centering
    \includegraphics[width=0.49\textwidth]{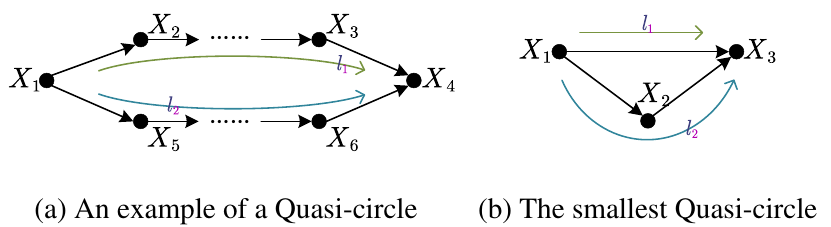}
    \caption{Some examples of quasi-circles.}
    \label{quasi-circle}
\end{figure}

Fig. \ref{quasi-circle} provides some examples of a quasi-circle. Specifically, Fig. \ref{quasi-circle}(a) illustrates a general quasi-circle, while Fig. \ref{quasi-circle}(b) shows the smallest quasi-circle. 
In this paper, we use the symbol $QC$ to denote a quasi-circle. For instance, $QC(X,Y,Z;X,W,Z)$ represents a quasi-circle composed of nodes $X, Y, Z, W$, and the paths $(X,Y,Z)$ and $(X,W,Z)$. Specifically, $QC(X, Y, Z)$ denotes a quasi-circle that contains only three nodes $X, Y, Z$.

\subsection{Reversed Direct Priors Lead to Quasi-circles}
\label{section_Reversed_Direct_Priors}

In this subsection, the relationship between quasi-circles and reversed direct prior is discussed to illustrate its effect on CSL.

\textbf{Lemma 1 (The reversed direct prior lead to the emergence of quasi-circles).} Assume that prior $e=(Y, X)$ is a reversed direct prior related to real edge $e'=(X, Y)$ in actual graph $G_0=(\mathbf{E},\mathbf{V})$. There are enough samples in the dataset $D$. Let $G_1$ denote the final result of a score-based methods, which employs a locally consistent scoring function $S(\cdot,\cdot)$ and accepts prior. If one of the following conditions satisfies:
\begin{equation}
\label{condition1}
\begin{aligned}
    1.&  \quad \exists \text{ node } Z\in \mathbf{V} \text{ s.t. } Z \in Pa^{G_0}(X)\cap Pa^{G_1}(X) \\
    2.&  \quad \exists \text{ node } Z\in \mathbf{V} \text{ s.t. } Z \in Pa^{G_0}(Y)\cap Pa^{G_1}(Y) \\
\end{aligned}    
\end{equation}
Then in $G_1$, there are edges between any two nodes among the three nodes $X$, $Y$, $Z$, and these edges form a quasi-circle, i.e.:
\begin{equation}
    QC(X,Y,Z) \in G_1
\end{equation}

\textbf{Proof of lemma 1.} This paper discusses separately according to whether $Z$ is the parent of $X$ or the parent of $Y$.

\begin{figure}[htbp]
    \centering
    \includegraphics[width=0.49\textwidth]{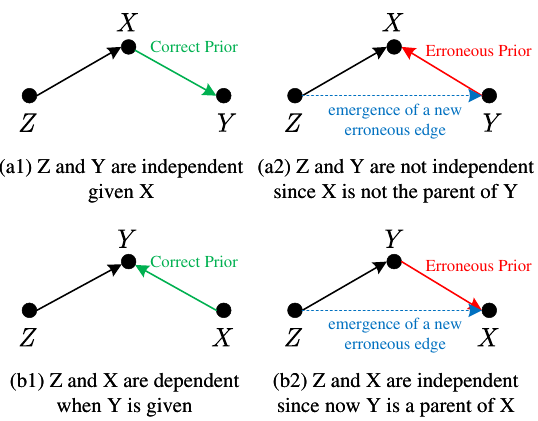}
    \caption{Schematic diagram of Lemma 1.}
    \label{lemma1}
\end{figure}

In the Case 1, $Z \in Pa^{G_0}(X)$ in the actual graph, as shown in Fig. \ref{lemma1}(a1). Due to $X \in Pa^{G_0}(Y)$ and $Z \in Pa^{G_0}(X)$, it could be derived that $Y\Vbar Z| Pa^{G_0}(Y)$. However, the situation is different in the final result graph $G_1$ who utilizes prior. Assume that there is no edge between $Z$ and $Y$ in $G_1$. Since the real edge $(X, Y)$ is replaced by the reversed direct prior $(Y, X)$, $Y \nVbar Z|Pa^{G_1}(Y)$, As shown in Fig. \ref{lemma1}(a2). According to local consistency, $S(G_1\cup (Y,Z),D)>S(G_1,D)$. This states that the edge $(Y,Z)$ should be added, contradicting the assumption that $G_1$ is the final result. Therefore, $G_1$ contains the edge connecting edges $Y$ and $Z$. 

For the Case 2, $Z \in Pa^{G_0}(Y)$. Assuming that there is no edge between $X$ and $Z$ in $G_0$, the real relationship of the three nodes is shown in Fig. \ref{lemma1}(b1). In $G_1$, $Y \in Pa^{G_1}(X)$, indicating that $X \nVbar Z |Pa^{G_1}(X)$ since $X$, $Y$, $Z$ form a collision structure in $G_0$, as shown in Fig. \ref{lemma1}(b2). Therefore, $S(G_1\cup (X,Z),D)>S(G_1,D)$, which contradicts the assumption. To sum up, there are edges between any two nodes among the three nodes $Y$, $Z$, and $X$, and these edges naturally form the quasi-circle $QC(X,Y,Z)$.

Lemma 1 shows that if either the head or tail node of the reversed direct prior  possesses the correct parent, the result graph will incorporate a quasi-circle that includes the this prior. Furthermore, if either the head or tail nodes have multiple correct parents, Lemma 1 suggests that multiple quasi-circles will form.

Some CSL methods will try to remove edges to improve the score. However, none of the edges of the generated quasi-circle will be removed in any edge deletion operation. This conclusion can also be substantiated by local consistency. 
First, Lemma 1 demonstrates that deleting the erroneous edges ($(Z,Y)$ in case 1 and $(Z,X)$ in case 2) decreases the score. For the remaining two edges, the skeletons associated with these edges are present in the actual graph, so according to local consistency, deleting any edge will diminish the graph's score. Hence, to maximize the graph's score, the quasi-circle will remain intact. This conclusion can further be employed to derive the following useful formula:
\begin{equation}
\label{euqation3}
    N_{QC^3}^{G_1}(e) > N_{QC^3}^{G_0}(\emptyset)
\end{equation}
where $e$ is a reversed direct prior, $N_{QC^3}^{G_1}(e)$ represents the number of quasi-circles with the prior $e$ and length 3 in the graph $G_1$ and $N_{QC^3}^{G_0}(\emptyset)$ represents corresponding number without prior in $G_0$. 

\subsection{Reversed Indirect Prior Lead to Quasi-circles}
Reversed indirect priors will also generate quasi-circles in the DAG obtained by the score-based method. More specifically, the following lemma will be proved:

\textbf{Lemma 2 (The reversed indirect prior lead to the emergence of quasi-circles).} Assume that prior $e=(Y, X)$ is a reversed indirect prior. $l_0=(X, Z,...,W,Y)$ is a real path in the actual graph $G_0=(E,V)$. There are enough samples in the dataset $D$. Let $G_1$ be the final result obtained by a score-based method employing a locally consistent scoring function $S(\cdot,\cdot)$ and accepts prior. Then $G_1$ contains at least one quasi-circle whose edges contain $e$ or edge in $l_0$.

\begin{figure}[htbp]
    \centering
    \includegraphics[width=0.49\textwidth]{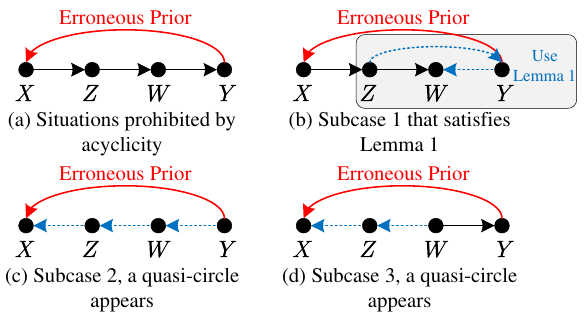}
    \caption{Schematic diagram of Lemma 2.}
    \label{lemma2}
\end{figure}

\textbf{Proof of Lemma 2.} According to local consistency, for each edge in $l_0$, the corresponding skeleton can be identified in $G_1$ using the score-based method, regardless of the usage of reversed indirect prior $e = (Y, X)$. However, reversed indirect prior $e$ influences the orientation of such a skeleton. In the following proof, let $l_1$ represent the path in $G_1$ that consists of these skeletons, which may not be directed. If every edge in $l_1$ aligns in direction with $l_0$, then these edges, alongside the reversed indirect prior $e$, will form a circle. The existence of a circle contradicts the definition of a DAG, hence a candidate graph with correct orientation for every edge is prohibited, as depicted in Fig. \ref{lemma2}(a) \footnote{In the proof of Lemma 2, $l_0=(X, Z, W, Y)$ is used as an example. This proof applies similarly for longer $l_0$.}. Thus, at least one edge in $l_1$ must have an orientation opposite to its corresponding edge on $l_0$.

To demonstrate that quasi-circles always exist, we divide the possible situations into three types for discussion, as shown in Fig. \ref{lemma2} (b, c, d).
In Fig. \ref{lemma2} (b), there are nodes $X$, $Y$ and $Z$ on $l_1$ that satisfy the conditions of Lemma 1, indicating that a quasi-circle of length 3 will be formed between these nodes. If there are no three nodes that satisfy the conditions of Lemma 1, the orientation of $l_1$ can only be Fig. \ref{lemma2} (c) or Fig. \ref{lemma2} (d). For Fig. \ref{lemma2} (c), $l_1$ and $e$ together form a quasi-circle. For Fig. \ref{lemma2}, all misoriented edges in $l_1$ constitute a directed path ending at $X$, while all correctly oriented edges, together with prior $e$, form a directed path ending at $X$. At this point, $l_1$ and the prior error $e=(Y, X)$ still form a quasi-circle. Consequently, $G_1$ will contain at least one quasi-circle, thus validating the conclusion.

Lemma 2 shows that if a prior is a reversed indirect prior, the result of the CSL will have quasi-circle containing misoriented edges. For a reversed indirect prior, Lemma 2 can be used for any path from the tail node to the head node. Therefore, the reversed indirect prior may yield multiple quasi-circles if there are multiple paths in the actual graph.

It is worth mentioning that if we assume the orientations of the edges in $l_1$ follow a binomial distribution, Fig. \ref{lemma2} (b) would have the highest probability of occurrence. Moreover, according to the conclusion of Lemma 1, the quasi-circle generated in Fig. \ref{lemma2} (b) will not be eliminated by any edge deletion operation. Consequently, among all the quasi-circles produced by the reversed indirect prior or reversed direct prior, the quasi-circles satisfying Lemma 1 might constitute the largest proportion. The length of this quasi-circle is 3. Thus, a quasi-circle of length 3 is a stable structure that can be used to signal incorrect edge orientation.

Lastly, this paper emphasizes that the foregoing discussion does not depend on any specific method. In fact, the proof of the conclusion primarily relies on the local consistency of the scoring function. As long as the score-based method employs a locally consistent scoring function, the above conclusions remain valid. More deeply, any score-based method searches for candidate graphs within a specific search space, where the search space typically lacks special properties. However, in the presence of priors, some candidate graphs might be unsearchable, which is equivalent to restricting the search of candidate graphs to a sub-search space. The distinctiveness of the sub-search space could affect the search process, and the graph with the maximum score might satisfy certain specific properties. This paper identifies a graph structure that must be present in the result graph, which could be beneficial for detecting potentially prior errors. Therefore, in Section \ref{section_algorithm}, this paper will leverage this graph structure to propose a strategy for detecting prior errors.

\subsection{The Effect of Indirect Priors and Irrelevant Priors}
We have proven that the reversed direct prior and reversed indirect prior leads to the generation of quasi-circles. In this subsection, we show that irrelevant and indirect priors do not significantly influence the results. In fact, irrelevant priors do not cause the learned graph to possess more missing, extra, or reversed edges, except themselves. More precisely, the following lemma is proposed:

\textbf{Lemma 3 (The influence of the irrelevant prior on  the learned DAG is only the addition of itself).} Assume that the prior $e$ is a irrelevant prior. Dataset $D$ contains enough samples. $G_0$ is the actual graph. $G_1$ is the results graph obtained by score-based methods, which employs a locally consistent scoring function and uses prior $e$ as hard constrains. Then we have:
\begin{equation}
    G_1=G_0 \cup e   
\end{equation}
where ``$=$'' indicates that two DAGs are equal in the sense of independency equivalence.

The concept of dependent equivalence mentioned in the above lemma is defined as follows:

\textbf{Definition (Independence Equivalent \cite{chickering2002optimal}):} Two DAGs $G$ and $H$ are independence equivalent if the independence constraints in the two DAGs are identical.

Independence equivalence describes the situation where different graphs may portray the same dependent or independent relationship between nodes. Since the correctness of the vast majority of score-based and constraint-based CSL methods relies on independence or dependencies between nodes, in many cases two independently equivalent graphs are indistinguishable. For Lemma 3, after adding priors, the causal learning process may change, which further leads to possible changes in the result graph. However, in addition to the influence brought by the irrelevant prior itself, the change of the result graph is limited to dependency equivalence, which is negligible in many tasks or applications. 

A similar conclusion holds for indirect priors:

\textbf{Lemma 4 (The influence of the indirect prior on the learned DAG is only the addition of itself).} Assume that the prior $e$ is a indirect prior. Dataset $D$ contains enough samples. $G_0$ is the actual graph. $G_1$ is the results graph obtained by score-based methods, which employs a locally consistent scoring function and uses prior $e$ as hard constrains. Then we have:
\begin{equation}
    G_1=G_0 \cup e   
\end{equation}

We focus more on the significance of Lemma 3 and Lemma 4 in the main text and place the proof in the appendix.
Lemma 3 and Lemma 4 demonstrate that indirect and irrelevant priors have less impact on the learned graph than order-reversed priors. Specifically, these two types of prior errors do not necessarily lead to quasi-circles. As a result, quasi-circles may not reliably indicate the presence of indirect and irrelevant priors. While this is regrettable, given the minor impact of these two types of priors, it is reasonable to assert that the detection of order-reversed priors is the focal point of this task.

\section{Quasi-circle-Based prior error Detection}
\label{section_algorithm}

The method proposed in this paper uses quasi-circles to detect prior errors. Intuitively, if a prior appears in the quasi-circles in the DAG obtained by the CSL method, the possibility that the prior is incorrect is greater. Therefore, the proposed method decides whether to suspect a prior based on whether the prior belongs to a quasi-circle. The specific process is shown in Alg. \ref{alg1}.

\begin{algorithm}[h]
\label{alg1}
\footnotesize
  \caption{Suspicious Prior Determination}
  \LinesNumbered
  \KwIn{The DAG $G$ obtained by the CSL method\; Prior set $P=\{e_1,e_2,...,e_n\}$\; Random variable set $\mathbf{X}=\{X_1,X_2,...,X_n\}$.}
  \KwOut{Suspicious prior set $Sus$.}
  \For{each prior $e=(X_i,X_j) \in P$}
  {
    $n_e=0$\\
    \For{$X_k \in X$}{
    \If{$X_i$, $X_j$, $X_k$ form a quasi-circle in $G$}
        {$n_e=n_e+1$}
    }
    \If{$n_e>0$}{
    Add $e$ to $Sus$
    }
  }
  \Return $Sus$
\end{algorithm}

Alg. \ref{alg1} suspects all prior that are contained in some quasi-circles. This setting may misjudge some correct priors related to quasi-circles that exist in actual graphs. Depending on the specific situation, a variety of options could also be used. For example, a threshold can be used to detect suspicious priors, or to consider a certain number of priors that occur most frequently in a quasi-circle as suspicious. 
The detection of suspicious priors primarily found quasi-circles of length 3 due to their high proportion proved in the Section \ref{section_method} and several considerations.
High algorithmic time complexity with increasing quasi-circle length in large node networks limits detection to shorter quasi-circles.
Also, from a counting point of view, a quasi-circle may have the same effect on the priors it contains, since each prior's count goes up by 1. Thus the longer the quasi-circle length, the less likely it will affect the relative order of the prior counts.

After detecting suspicious priors, the proposed algorithm improves the result of causal learning by updating the priors. If a prior $e=(X_i,X_j)$ is suspect, it will be replaced by another prior $Re(e)$, whose expression is represented as:
\begin{equation}
    Re(e)=(X_j,X_i), \forall  e=(X_i,X_j)
\end{equation}
After modifying the priors, the based CSL method is used again with the updated set of priors as input, resulting in improved causal learning results. The specific algorithm is shown in Alg \ref{alg2}.

\begin{algorithm}[h]
\label{alg2}
\footnotesize
  \caption{Prior Modification}
  \LinesNumbered
  \KwIn{Dataset $D$\; Prior set $P=\{e_1,e_2,...,e_n\}$\; Iteration limit $IL$.}
  \KwOut{DAG $G$.}
  Find $G$ using any CSL methods\\
  $It=0$\\
  Find suspicious prior $Sus$ using Alg.\ref{alg1}\\
  $Sus_{rec}=Sus$\\
  \While{$|Sus|!=0$ or $It < IL$}{
  $It=It+1$\\
  \For{$e \in P$}{
  \eIf{$e \in Sus$ and $Re(e) \notin Sus_{rec}$}
  {$P=P\cup \{Re(e)\}\textbackslash e$}
  {\If{$e \in Sus$ and $Re(e) \in Sus_{rec}$}{
  $P=P\textbackslash e$
  }}
  }
  Find $G$ using any CSL methods\\
  Find suspicious prior $Sus$ using Alg.\ref{alg1}\\
  $Sus_{rec}=Sus_{rec}\cup Sus$
  }
  \Return $G$
\end{algorithm}

During the actual implementation of the proposed algorithm, the input prior modification process terminates either when no new suspicious priors are detected or when the number of iterations reaches a predetermined value. The causal learning result based on the input priors set $P$ at that time is considered as the final result. Depending on the termination or suspicious prior assessment conditions, the modification process may occur in multiple rounds, and a prior may be suspected multiple times. For instance, in the first round, the prior $e$ may be deemed suspicious. In the second round, the reverse of $e$, denoted $Re(e)$, may also be considered suspect. If both a prior and its reverse are suspected, it implies that the corresponding two nodes might not have a causal relationship. In such a case, the proposed algorithm removes the prior $e$ and $Re(e)$ from the priors set and consistently excludes these priors from the set of candidate edges when adding edges.

\section{Experiments}
\label{section_experiment}

This section first introduces the basic settings of the experiment, and then shows the advantages of the QC-based prior correction strategy from different perspectives. The content mainly includes (1) The improvement of QC-based prior correction strategy on the based method, (2) Performance comparison of different CSL methods, (3) The resilience test of different methods to prior errors.
More experimental results can be found in the appendix, including (4) Experimental verification of the effect of four kinds of prior errors, (5) Comparison of the error correction capabilities of the proposed strategy and soft constraint methods, (6) More extended QC-based methods and their performance.
\subsection{Setup}
\label{section_setup}

\textbf{Methods Used.} This paper implements two QC-based prior correction method named HC-QC and NOTEARS-QC, which utilize Hill-Climbing (HC) and NOTEARS respectively as the foundational models within Alg. 2. These two methods are selected for they are representative in traditional and novel gradient-based methods, respectively. The comparative methods involved in the experiment include HC, MMHC\footnote{The codes of HC and MMHC can be found at https://github.com/Enderlogic/MMHC-Python}, PC\footnote{https://github.com/py-why/causal-learn}, CaMML\footnote{https://bayesian-intelligence.com/software/}, GranDAG \cite{lachapelle2019gradient}, GAE \cite{ng2019graph}, DAG-GNN \cite{yu2019dag}, GOLEM\footnote{The codes of NOTEARS, GranDAG, GAE, DAG-GNN, GOLEM can be found at https://github.com/huawei-noah/trustworthyAI/tree/master/gcastle}\cite{ng2020role}. Some of these methods have been introduced in Section \ref{section_Principal_CSL_Methods}, and the rest can be found in the appendix.

\textbf{Datasets Used.} The experiment covers real, pesudo-real and simulated data. Real data include SACHS (a well known data that quantifies the expression levels of various proteins and phospholipids in human cells) \cite{sachs2005causal}, ECOLI70 (data measuring the stress response of the microorganism Escherichia coli during expression of a recombinant protein) \cite{schafer2005shrinkage}, and MAGIC-NIAB (genotypes and traits on a winter wheat population produced by the UK National Institute of Agricultural Botany) \cite{scutari2014multiple}. For pesudo-real data, CHILD, ALARM, and INSURANCE accessed from the Bayesian network repository\footnote{https://www.bnlearn.com/} are utilized. The difference between pesudo-real data and real data lies in the fact that the former is generated based on true network structure, while the latter encompasses both genuine data and authentic network structure. The information of corresponding networks is shown in the Table \ref{graph_information}. For simulated data, a random DAG with a certain number of nodes and edges is generated using the Erdös-Rényi method \cite{martinez2020computational}. Each pesudo-real and simulated data consist of randomly drawn samples from the joint probability distribution of corresponding set of random variables. CHILD, ALARM, and INSURANCE dataset have 2000, 4000, 4000 samples, respectively, while for the simulated data, an appropriate capacity is selected according to the number of nodes in the network.

\begin{table}[htbp]
\centering
\caption{Information about the Used Networks}
\setlength{\tabcolsep}{1.3mm}{
\begin{tabular}{c|ccccc}
\hline
Network   & Nodes & Arcs & Parameters & Average Degree & Data Type\\ \hline
ASIA      & 8     & 8    & 18         & 2.00   &  Pseudo-Real       \\
ALARM     & 37    & 46   & 509        & 2.49   &Pseudo-Real        \\
CHILD     & 20    & 25   & 230        & 2.50   &Pseudo-Real        \\
INSURANCE & 27    & 52   & 1008       & 3.85   &Pseudo-Real        \\ 
SACHS & 11    & 17   & 178       & 3.09        &  Real \\
ECOLI70 & 46    & 70   & 162       & 3.04    &  Real       \\
MAGIC-NIAB & 44    & 66   & 154       & 3.00    &  Real       \\
\hline
\end{tabular}}
\label{graph_information}
\end{table}

\textbf{Parameter Settings} The parameter of the proposed strategy is only the iteration limit IL, which is set to 5. As for the parameters of the comparison methods, the default parameters of the source code are used.

\textbf{Selection and Usage of Prior.} The prior is selected with equal probability from the set of candidate edges. The set for the correct prior is the set of real edges, whereas the set for the prior errors may include the reverse of the true edge or node pairs that are not directly adjacent, depending on the type of prior error required. To satisfy the definition of a DAG, the priors are selected sequentially. A prior selected later cannot violate the priors selected earlier, nor can it form a circle with the existing priors.

\textbf{Evaluation Metric.} In the experiments, the Structural Hamming Distance (SHD) is utlized to evaluate the performance of CSL methods, which comprises three sub-items: the number of missing edges, extra edges, and reversed edges.
In order to facilitate the comparison of different methods, some other indicators are also calculated, including the increment of SHD after adding the prior, the retention rate of the correct prior, and the detection rate of the prior errors.

\textbf{Experimental Hardware.} All experiments are conducted with an Intel(R) Core (TM) i7-12700K CPU, 32GB RAM.

\subsection{The Improvement of QC-based Prior Correction Strategy on the Based Method.}
\label{section_improvement_based_method}
\begin{table*}[htbp]
\centering
\setlength{\tabcolsep}{1.3mm}
\caption{The Performance of the HC-QC and HC on Different Datasets under Different Prior Settings}
\begin{tabular}{c|c||cccc||cccccccc}
\hline
\multirow{2}{*}{Network}   & \multirow{2}{*}{\begin{tabular}[c]{@{}c@{}}Number of Prior \\ (right, wrong)\end{tabular}} & \multicolumn{4}{c||}{HC}                                                                           & \multicolumn{8}{c}{HC-QC}                                                                                                               \\ \cline{3-14} 
                           &                                                                                  & \multicolumn{1}{c|}{Missing} & \multicolumn{1}{c|}{Extra} & \multicolumn{1}{c|}{Reversed} & SHD   & \multicolumn{2}{c|}{Missing}        & \multicolumn{2}{c|}{Extra}          & \multicolumn{2}{c|}{Reversed}        & \multicolumn{2}{c}{SHD} \\ \hline\hline
\multirow{9}{*}{CHILD}     & (0,0)                                                                            & \multicolumn{1}{c|}{1}       & \multicolumn{1}{c|}{3}     & \multicolumn{1}{c|}{8}        & 12    & 1     & \multicolumn{1}{c|}{0\%}     & 3    & \multicolumn{1}{c|}{0\%}     & 8     & \multicolumn{1}{c|}{0\%}     & 12        & 0\%         \\
                           & (0,2)                                                                            & \multicolumn{1}{c|}{1.82}    & \multicolumn{1}{c|}{3.39}  & \multicolumn{1}{c|}{8.41}     & 13.62 & 1.84  & \multicolumn{1}{c|}{1.1\%}   & 2.29 & \multicolumn{1}{c|}{-32.4\%} & 7.51  & \multicolumn{1}{c|}{-10.7\%} & 11.64     & \textbf{-14.5\%}     \\
                           & (0,4)                                                                            & \multicolumn{1}{c|}{2.13}    & \multicolumn{1}{c|}{4.06}  & \multicolumn{1}{c|}{9.53}     & 15.72 & 1.82  & \multicolumn{1}{c|}{-14.6\%} & 2.21 & \multicolumn{1}{c|}{-45.6\%} & 7.82  & \multicolumn{1}{c|}{-17.9\%} & 11.85     & \textbf{-24.6\%}     \\
                           & (2,0)                                                                            & \multicolumn{1}{c|}{0.98}    & \multicolumn{1}{c|}{2.08}  & \multicolumn{1}{c|}{7.10}      & 10.16 & 1.05  & \multicolumn{1}{c|}{7.1\%}   & 2.08 & \multicolumn{1}{c|}{0\%}     & 7.22  & \multicolumn{1}{c|}{1.7\%}   & 10.35     & 1.9\%       \\
                           & (2,2)                                                                            & \multicolumn{1}{c|}{1.68}    & \multicolumn{1}{c|}{2.73}  & \multicolumn{1}{c|}{7.56}     & 11.97 & 1.61  & \multicolumn{1}{c|}{-4.2\%}  & 1.79 & \multicolumn{1}{c|}{-34.4\%} & 6.87  & \multicolumn{1}{c|}{-9.1\%}  & 10.27     & \textbf{-14.2\%}     \\
                           & (2,4)                                                                            & \multicolumn{1}{c|}{1.76}    & \multicolumn{1}{c|}{3.48}  & \multicolumn{1}{c|}{8.80}      & 14.04 & 1.58  & \multicolumn{1}{c|}{-10.2\%} & 1.90  & \multicolumn{1}{c|}{-45.4\%} & 7.40   & \multicolumn{1}{c|}{-15.9\%} & 10.88     & \textbf{-22.5\%}     \\
                           & (4,0)                                                                            & \multicolumn{1}{c|}{1.01}    & \multicolumn{1}{c|}{1.43}  & \multicolumn{1}{c|}{6.11}     & 8.55  & 1.14  & \multicolumn{1}{c|}{12.9\%}  & 1.41 & \multicolumn{1}{c|}{-1.4\%}  & 6.35  & \multicolumn{1}{c|}{3.9\%}   & 8.90       & 4.1\%       \\
                           & (4,2)                                                                            & \multicolumn{1}{c|}{1.38}    & \multicolumn{1}{c|}{2.12}  & \multicolumn{1}{c|}{6.77}     & 10.27 & 1.43  & \multicolumn{1}{c|}{3.6\%}   & 1.12 & \multicolumn{1}{c|}{-47.2\%} & 5.97  & \multicolumn{1}{c|}{-11.8\%} & 8.52      & \textbf{-17\%}       \\
                           & (4,4)                                                                            & \multicolumn{1}{c|}{1.57}    & \multicolumn{1}{c|}{3.04}  & \multicolumn{1}{c|}{8.06}     & 12.67 & 1.53  & \multicolumn{1}{c|}{-2.5\%}  & 1.71 & \multicolumn{1}{c|}{-43.8\%} & 6.67  & \multicolumn{1}{c|}{-17.2\%} & 9.91      & \textbf{-21.8\%}     \\ \hline\hline
\multirow{9}{*}{ALARM}     & (0,0)                                                                            & \multicolumn{1}{c|}{5}       & \multicolumn{1}{c|}{5}     & \multicolumn{1}{c|}{13}       & 23    & 5     & \multicolumn{1}{c|}{0\%}    & 5    & \multicolumn{1}{c|}{0\%}     & 13    & \multicolumn{1}{c|}{0\%}     & 23        & 0\%         \\
                           & (0,2)                                                                            & \multicolumn{1}{c|}{4.92}    & \multicolumn{1}{c|}{6.84}  & \multicolumn{1}{c|}{15.19}    & 26.95 & 4.73  & \multicolumn{1}{c|}{-3.9\%} & 5.13 & \multicolumn{1}{c|}{-25\%}   & 13.18 & \multicolumn{1}{c|}{-13.2\%} & 23.04     & \textbf{-14.5\%}     \\
                           & (0,4)                                                                            & \multicolumn{1}{c|}{5.09}    & \multicolumn{1}{c|}{8.56}  & \multicolumn{1}{c|}{17.23}    & 30.88 & 4.58  & \multicolumn{1}{c|}{-10\%}  & 5.18 & \multicolumn{1}{c|}{-39.5\%} & 13.17 & \multicolumn{1}{c|}{-23.6\%} & 22.93     & \textbf{-25.7\%}     \\
                           & (2,0)                                                                            & \multicolumn{1}{c|}{4.78}    & \multicolumn{1}{c|}{4.55}  & \multicolumn{1}{c|}{11.54}    & 20.87 & 4.86  & \multicolumn{1}{c|}{1.7\%}  & 4.66 & \multicolumn{1}{c|}{2.4\%}   & 11.81 & \multicolumn{1}{c|}{2.3\%}   & 21.33     & 2.2\%       \\
                           & (2,2)                                                                            & \multicolumn{1}{c|}{4.72}    & \multicolumn{1}{c|}{6.54}  & \multicolumn{1}{c|}{13.75}    & 25.01 & 4.62  & \multicolumn{1}{c|}{-2.1\%} & 4.97 & \multicolumn{1}{c|}{-24\%}   & 12.22 & \multicolumn{1}{c|}{-11.1\%} & 21.81     & \textbf{-12.8\%}     \\
                           & (2,4)                                                                            & \multicolumn{1}{c|}{4.71}    & \multicolumn{1}{c|}{8.10}   & \multicolumn{1}{c|}{16.06}    & 28.87 & 4.44  & \multicolumn{1}{c|}{-5.7\%} & 5.08 & \multicolumn{1}{c|}{-37.3\%} & 12.37 & \multicolumn{1}{c|}{-23\%}   & 21.89     & \textbf{-24.2\%}     \\
                           & (4,0)                                                                            & \multicolumn{1}{c|}{4.47}    & \multicolumn{1}{c|}{4.41}  & \multicolumn{1}{c|}{10.53}    & 19.41 & 4.66  & \multicolumn{1}{c|}{4.3\%}  & 4.69 & \multicolumn{1}{c|}{6.3\%}   & 11.25 & \multicolumn{1}{c|}{6.8\%}   & 20.60      & 6.1\%       \\
                           & (4,2)                                                                            & \multicolumn{1}{c|}{4.40}     & \multicolumn{1}{c|}{6.40}   & \multicolumn{1}{c|}{13.08}    & 23.88 & 4.42  & \multicolumn{1}{c|}{0.5\%}  & 5.18 & \multicolumn{1}{c|}{-19.1\%} & 12.09 & \multicolumn{1}{c|}{-7.6\%}  & 21.69     & \textbf{-9.2\%}      \\
                           & (4,4)                                                                            & \multicolumn{1}{c|}{4.33}    & \multicolumn{1}{c|}{8.00}     & \multicolumn{1}{c|}{15.07}    & 27.40  & 4.25  & \multicolumn{1}{c|}{-1.8\%} & 5.24 & \multicolumn{1}{c|}{-34.5\%} & 12.03 & \multicolumn{1}{c|}{-20.2\%} & 21.52     & \textbf{-21.5\%}     \\ \hline\hline
\multirow{9}{*}{INSURANCE} & (0,0)                                                                            & \multicolumn{1}{c|}{15}      & \multicolumn{1}{c|}{11}    & \multicolumn{1}{c|}{13}       & 39    & 15    & \multicolumn{1}{c|}{0\%}    & 11   & \multicolumn{1}{c|}{0\%}     & 13    & \multicolumn{1}{c|}{0\%}     & 39        & 0\%         \\
                           & (0,2)                                                                            & \multicolumn{1}{c|}{13.57}   & \multicolumn{1}{c|}{9.62}  & \multicolumn{1}{c|}{14.21}    & 37.40  & 13.30  & \multicolumn{1}{c|}{-2\%}   & 8.74 & \multicolumn{1}{c|}{-9.1\%}  & 12.30  & \multicolumn{1}{c|}{-13.4\%} & 34.34     & \textbf{-8.2\%}      \\
                           & (0,4)                                                                            & \multicolumn{1}{c|}{12.88}   & \multicolumn{1}{c|}{9.54}  & \multicolumn{1}{c|}{15.10}     & 37.52 & 12.51 & \multicolumn{1}{c|}{-2.9\%} & 7.89 & \multicolumn{1}{c|}{-17.3\%} & 12.00    & \multicolumn{1}{c|}{-20.5\%} & 32.40      & \textbf{-13.6\% }    \\
                           & (2,0)                                                                            & \multicolumn{1}{c|}{13.38}   & \multicolumn{1}{c|}{9.04}  & \multicolumn{1}{c|}{12.11}    & 34.53 & 13.49 & \multicolumn{1}{c|}{0.8\%}  & 8.74 & \multicolumn{1}{c|}{-3.3\%}  & 12.08 & \multicolumn{1}{c|}{-0.2\%}  & 34.31     & -0.6\%      \\
                           & (2,2)                                                                            & \multicolumn{1}{c|}{12.70}    & \multicolumn{1}{c|}{8.60}   & \multicolumn{1}{c|}{13.07}    & 34.37 & 12.23 & \multicolumn{1}{c|}{-3.7\%} & 7.32 & \multicolumn{1}{c|}{-14.9\%} & 11.34 & \multicolumn{1}{c|}{-13.2\%} & 30.89     & \textbf{-10.1\%}     \\
                           & (2,4)                                                                            & \multicolumn{1}{c|}{12.35}   & \multicolumn{1}{c|}{9.19}  & \multicolumn{1}{c|}{14.45}    & 35.99 & 11.59 & \multicolumn{1}{c|}{-6.2\%} & 7.05 & \multicolumn{1}{c|}{-23.3\%} & 11.71 & \multicolumn{1}{c|}{-19\%}   & 30.35     & \textbf{-15.7\%}    \\
                           & (4,0)                                                                            & \multicolumn{1}{c|}{12.35}   & \multicolumn{1}{c|}{7.77}  & \multicolumn{1}{c|}{10.80}     & 30.92 & 12.57 & \multicolumn{1}{c|}{1.8\%}  & 7.40  & \multicolumn{1}{c|}{-4.8\%}  & 10.63 & \multicolumn{1}{c|}{-1.6\%}  & 30.60      & -1\%        \\
                           & (4,2)                                                                            & \multicolumn{1}{c|}{11.72}   & \multicolumn{1}{c|}{7.53}  & \multicolumn{1}{c|}{11.76}    & 31.01 & 11.32 & \multicolumn{1}{c|}{-3.4\%} & 6.66 & \multicolumn{1}{c|}{-11.6\%} & 10.72 & \multicolumn{1}{c|}{-8.8\%}  & 28.70      & \textbf{-7.4\%}      \\
                           & (4,4)                                                                            & \multicolumn{1}{c|}{11.47}   & \multicolumn{1}{c|}{8.35}  & \multicolumn{1}{c|}{13.07}    & 32.89 & 10.70  & \multicolumn{1}{c|}{-6.7\%} & 6.33 & \multicolumn{1}{c|}{-24.2\%} & 11.01 & \multicolumn{1}{c|}{-15.8\%} & 28.04     & \textbf{-14.7\%}     \\ \hline
\end{tabular}
\label{ComparedWithHC}
\end{table*}

In this subsection, the proposed method is compared with the basic method to demonstrate the advantage of the ability to identify prior errors.

Experiments on HC-QC are presented first. The used priors sets are characterized by two parameters: the number of correct priors and the number of prior errors, which are taken as 0, 2, and 4, respectively. Each experiment randomly selects correct and erroneous priors based on the real network structure. For each parameter setting, the experiment is performed 100 times. The results are shown in Table \ref{ComparedWithHC}. In addition to SHD, the magnitude of the improvement of HC-QC over SHD is also reported. It can be seen that in all datasets, as the number of prior errors increases, the ability of HC methods to learn causal structure decreases gradually, even though correct priors dominate the set of priors. Although HC-QC is also disturbed by prior errors, the performance of HC-QC is comparatively better than HC in most scenarios, indicating that HC-QC is effective. The performance improvement of HC-QC is mainly due to the reduction in the number of extras and reversed edges, in which the number of reversed edges is likely to be lower, and the number of extras can be reduced by more than 10\% on average. Combined with the properties of different prior errors, the above phenomena suggests that HC-QC can remove or correct reversed direct priors or reversed indirect priors. 

The larger the number of prior errors, the stronger the performance improvement of the HC-QC. While HC-QC may lead to slight performance degradation when there are no or only a few prior errors in the priors set. This is mainly due to the fact that the main idea of the proposed method is to judge prior errors based on the occurrence of quasi-circles, so it is inevitable to mistakenly exclude some correct priors, especially when in the case that the real network itself contains quasi-circles and its edges are selected as prior at the same time. In real scenarios, the prior may not be completely reliable. In this case, HC-QC could improve the quality of the priors sets, which is an advantage over HC method that directly use the prior without suspicion.

\begin{table*}[ht]
\centering
\caption{Performance of Quasi-Circles-based Prior Correction Strategy  strategies in irrelevant prior scenarios}
\begin{tabular}{ccc|cc|cc|cc|cc}
\multicolumn{3}{c|}{Prior   Proportion}                                                                                                                                       & \multicolumn{2}{c|}{10\%}                                  & \multicolumn{2}{c|}{20\%}                                  & \multicolumn{2}{c|}{30\%}                                  & \multicolumn{2}{c}{40\%}                                   \\ \hline
\multicolumn{3}{c|}{Metric}                                                                                                                                                   & F1                           & SHD                        & F1                           & SHD                        & F1                           & SHD                        & F1                           & SHD                        \\ \hline
\multicolumn{1}{c|}{}                             & \multicolumn{1}{c|}{}                                                                                           & NOTEARS & 30.3                         & 20                         & \cellcolor[HTML]{D9D9D9}16.7 & \cellcolor[HTML]{D9D9D9}27 & \cellcolor[HTML]{D9D9D9}27   & \cellcolor[HTML]{D9D9D9}24 & \cellcolor[HTML]{D9D9D9}27.9 & \cellcolor[HTML]{D9D9D9}26 \\
\multicolumn{1}{c|}{}                             & \multicolumn{1}{c|}{\multirow{-2}{*}{\begin{tabular}[c]{@{}c@{}}Prior Error \\      Ratio =0.5\end{tabular}}}   & +QC     & 30.3                         & 20                         & \cellcolor[HTML]{D9D9D9}20.8 & \cellcolor[HTML]{D9D9D9}20 & \cellcolor[HTML]{D9D9D9}35.3 & \cellcolor[HTML]{D9D9D9}19 & \cellcolor[HTML]{D9D9D9}30.3 & \cellcolor[HTML]{D9D9D9}20 \\ \cline{2-11} 
\multicolumn{1}{c|}{}                             & \multicolumn{1}{c|}{}                                                                                           & NOTEARS & \cellcolor[HTML]{D9D9D9}30.3 & \cellcolor[HTML]{D9D9D9}20 & \cellcolor[HTML]{D9D9D9}16.7 & \cellcolor[HTML]{D9D9D9}27 & \cellcolor[HTML]{D9D9D9}26.3 & \cellcolor[HTML]{D9D9D9}22 & \cellcolor[HTML]{D9D9D9}27.3 & \cellcolor[HTML]{D9D9D9}26 \\
\multicolumn{1}{c|}{\multirow{-4}{*}{SACHS}}      & \multicolumn{1}{c|}{\multirow{-2}{*}{\begin{tabular}[c]{@{}c@{}}Prior Error \\      Ratio =1.0\end{tabular}}} & +QC     & \cellcolor[HTML]{D9D9D9}32.1 & \cellcolor[HTML]{D9D9D9}19 & \cellcolor[HTML]{D9D9D9}18.8 & \cellcolor[HTML]{D9D9D9}21 & \cellcolor[HTML]{D9D9D9}38.9 & \cellcolor[HTML]{D9D9D9}21 & \cellcolor[HTML]{D9D9D9}34.3 & \cellcolor[HTML]{D9D9D9}20 \\ \hline
\multicolumn{1}{c|}{}                             & \multicolumn{1}{c|}{}                                                                                           & NOTEARS & \cellcolor[HTML]{D9D9D9}15.8 & \cellcolor[HTML]{D9D9D9}78 & \cellcolor[HTML]{F2F2F2}27.8 & \cellcolor[HTML]{F2F2F2}72 & \cellcolor[HTML]{D9D9D9}30.5 & \cellcolor[HTML]{D9D9D9}70 & \cellcolor[HTML]{D9D9D9}40.7 & \cellcolor[HTML]{D9D9D9}61 \\
\multicolumn{1}{c|}{}                             & \multicolumn{1}{c|}{\multirow{-2}{*}{\begin{tabular}[c]{@{}c@{}}Prior Error\\      Ratio =0.5\end{tabular}}}   & +QC     & \cellcolor[HTML]{D9D9D9}18.9 & \cellcolor[HTML]{D9D9D9}73 & \cellcolor[HTML]{F2F2F2}19.6 & \cellcolor[HTML]{F2F2F2}76 & \cellcolor[HTML]{D9D9D9}32.4 & \cellcolor[HTML]{D9D9D9}69 & \cellcolor[HTML]{D9D9D9}42.9 & \cellcolor[HTML]{D9D9D9}57 \\ \cline{2-11} 
\multicolumn{1}{c|}{}                             & \multicolumn{1}{c|}{}                                                                                           & NOTEARS & \cellcolor[HTML]{D9D9D9}15.8 & \cellcolor[HTML]{D9D9D9}78 & \cellcolor[HTML]{D9D9D9}24.1 & \cellcolor[HTML]{D9D9D9}72 & 29.8                         & 66                         & \cellcolor[HTML]{D9D9D9}35.4 & \cellcolor[HTML]{D9D9D9}64 \\
\multicolumn{1}{c|}{\multirow{-4}{*}{MAGIC-NIAB}} & \multicolumn{1}{c|}{\multirow{-2}{*}{\begin{tabular}[c]{@{}c@{}}Prior Error\\      Ratio =1.0\end{tabular}}} & +QC     & \cellcolor[HTML]{D9D9D9}18.9 & \cellcolor[HTML]{D9D9D9}73 & \cellcolor[HTML]{D9D9D9}27.8 & \cellcolor[HTML]{D9D9D9}69 & 26.8                         & 70                         & \cellcolor[HTML]{D9D9D9}36.7 & \cellcolor[HTML]{D9D9D9}63 \\ \hline
\multicolumn{1}{c|}{}                             & \multicolumn{1}{c|}{}                                                                                           & NOTEARS & \cellcolor[HTML]{D9D9D9}17.6 & \cellcolor[HTML]{D9D9D9}91 & 31.9                         & 81                         & \cellcolor[HTML]{D9D9D9}36.7 & \cellcolor[HTML]{D9D9D9}77 & \cellcolor[HTML]{D9D9D9}37   & \cellcolor[HTML]{D9D9D9}75 \\
\multicolumn{1}{c|}{}                             & \multicolumn{1}{c|}{\multirow{-2}{*}{\begin{tabular}[c]{@{}c@{}}Prior Error\\      Ratio =0.5\end{tabular}}}   & +QC     & \cellcolor[HTML]{D9D9D9}22.2 & \cellcolor[HTML]{D9D9D9}81 & 25.4                         & 83                         & \cellcolor[HTML]{D9D9D9}37.6 & \cellcolor[HTML]{D9D9D9}68 & \cellcolor[HTML]{D9D9D9}41.6 & \cellcolor[HTML]{D9D9D9}74 \\ \cline{2-11} 
\multicolumn{1}{c|}{}                             & \multicolumn{1}{c|}{}                                                                                           & NOTEARS & \cellcolor[HTML]{D9D9D9}21.5 & \cellcolor[HTML]{D9D9D9}89 & \cellcolor[HTML]{D9D9D9}25.4 & \cellcolor[HTML]{D9D9D9}82 & \cellcolor[HTML]{F2F2F2}33.1 & \cellcolor[HTML]{D9D9D9}78 & \cellcolor[HTML]{D9D9D9}29.9 & \cellcolor[HTML]{D9D9D9}84 \\
\multicolumn{1}{c|}{\multirow{-4}{*}{ECOLI70}}      & \multicolumn{1}{c|}{\multirow{-2}{*}{\begin{tabular}[c]{@{}c@{}}Prior Error\\      Ratio =1.0\end{tabular}}} & +QC     & \cellcolor[HTML]{D9D9D9}23.3 & \cellcolor[HTML]{D9D9D9}80 & \cellcolor[HTML]{D9D9D9}30.8 & \cellcolor[HTML]{D9D9D9}81 & \cellcolor[HTML]{F2F2F2}31.6 & \cellcolor[HTML]{D9D9D9}74 & \cellcolor[HTML]{D9D9D9}36.4 & \cellcolor[HTML]{D9D9D9}82
\end{tabular}
\begin{flushleft}
``+QC" indicates NOTEARS-QC which utilizes proposed prior correction strategy.
\end{flushleft}
\label{realdata}
\end{table*}

The following experiments are conducted on real data. The comparative result of NOTEARS-QC and NOTEARS, outlined in Table \ref{realdata}, reveal that NOTEARS-QC consistently outperforms NOTEARS across a majority of test scenarios. This finding is consistent with the results of HC-QC. It's noteworthy that the performance improvements on real datasets, compared with Table \ref{ComparedWithHC}, are relatively poor. This discrepancy underscores the inherent challenges associated with causal structure discovery in real-world data, which may be attributed to various factors including limited sample sizes, the presence of noise and sampling bias, heterogeneity in data types (continuous and discrete variables), and instances of missing data. Despite these limitations, the overall superior performance of NOTEARS-QC substantiates the efficacy of our proposed approach, indicating its potential to advance the field of causal structure discovery, especially in challenging real-world settings.

\subsection{Performance Comparison of Different CSL Methods}
\label{section_comparison_Different_Methods}
In this subsection, we compare the proposed method with other methods. The comparison centers on HC-QC and NOTEARS-QC, respectively. The former HC-QC will be compared with classic methods of various branches to illustrate the generality of prior errors identification problems. The latter NOTEARS-QC will be compared with NOTEARS and other cutting-edge methods to prove the superior performance of this strategy.

\subsubsection{Comparison with classic methods}
\label{section_comparison_classic_Methods}
\begin{table*}[htbp]
\centering
\caption{Performance Changes of Different Causal Structure Learning Methods in the face of Prior Errors}
\begin{tabular}{c|c||c||c|c|c}
\hline
\multirow{2}{*}{Network}   & \multirow{2}{*}{Method} & \multicolumn{4}{c}{Number of  Prior Errors}                                                                                 \\ \cline{3-6} 
                           &                         & 0                            & 2                             & 4                              & 6                              \\ \hline\hline
\multirow{5}{*}{Child}     & HC                      & 1 / 3 / 8 / 12               & +0.82 / +0.39 / +0.41 / +1.62 & +1.13 / +1.06 / +1.53 / +3.72  & +1.31 / +1.79 / +2.73 / +5.83  \\
                           & PC                      & 2 / 0 / 6 / 8            & -0.12 / +0 / +2.15 / +2.03    & -0.25 / +0 / +4.16 / +3.91   & -0.37 / +0 / +6.04 / +5.67     \\
                           & MMHC                    & 6 / 0 / 6 / 12               & -0.43 / +0 / +1.02 / +0.59    & -0.86 / +0 / +2.25 / +1.39     & -1.24 / +0 / +3.61 / +2.37     \\
                           & CAMML                   & 11.31 / 6.41 / 0.59 / 18.31              & +0.01 / -0.11 / +1.40 / +1.30       & +0 / -0.04 / +2.16 / +2.12      & +0 / -0.03 / +2.35 / +2.32      \\
                           & HC-QC                & 1 / 3 / 8 / 12               & +0.84 / -0.71 / -0.49 / \textbf{-0.36} & +0.82 / -0.79 / -0.18 / \textbf{-0.15}  & +0.96 / -0.42 / +0.36 / \textbf{+0.90}   \\ \hline\hline
\multirow{5}{*}{Alarm}     & HC                      & 5 / 5 / 13 / 23              & -0.08 / +1.84 / +2.19 / +3.95 & +0.09 / +3.56 / +4.23 / +7.88  & +0.06 / +4.80 / +6.07 / +10.93  \\
                           & PC                      & 4 / 1 / 4 / 9                & -0.25 / +0 / +2.53 / +2.28    & -0.39 / +0 / +5.00 / +4.61        & -0.52 / +0 / +7.39 / +6.87   \\
                           & MMHC                    & 11 / 0 / 5 / 16              & -0.38 / +0.14 / +2.91 / +2.67 & -0.78 / +0.30 / +5.44 / +4.96   & -1.21 / +0.42 / +8.03 / +7.24  \\
                           & CAMML                   & 17.64 / 16.05 / 4.67 / 38.36 & -0.10 / +0.33 / +1.07 / +1.30   & -0.14 / +0.46 / +2.11 / +2.43    & -0.05 / +0.47 / +2.76 / +3.18  \\
                           & HC-QC                & 5 / 5 / 13 / 23              & -0.27 / +0.13 / +0.18 / \textbf{+0.04} & -0.42 / +0.18 / +0.17 / \textbf{-0.07}  & -0.55 / +0.34 / +0.16 / \textbf{-0.05}  \\ \hline\hline
\multirow{5}{*}{Insurance} & HC                      & 15 / 11 / 13 / 39            & -1.43 / -1.38 / +1.21 / -1.60  & -2.12 / -1.46 / +2.10 / -1.48   & -2.57 / -0.85 / +3.48 / +0.06  \\
                           & PC                      & 9 / 2 / 11 / 22          & -0.24 / -0.08 / +1.84 / +1.52 & -0.53 / -0.14 / +3.32 / +2.65  & -0.90 / -0.22 / +5.06 / +3.94 \\
                           & MMHC                    & 22 / 2 / 11 / 35             & -0.86 / +0.12 / +0.43 / -0.31 & -1.61 / +0.20 / +1.21 / -0.20   & -2.41 / +0.31 / +2.53 / +0.43  \\
                           & CAMML                   & 21.12 / 5.43 / 9.10 / 35.65        & +0 / +0.05 / -0.48 / -0.43     & +0 / +0.18 / -0.30 / -0.12      & +0 / +0.16 / +0.19 / +0.35      \\
                           & HC-QC                & 15 / 11 / 13 / 39            & -1.70 / -2.26 / -0.70 / \textbf{-4.66}  & -2.49 / -3.11 / -1.00 / \textbf{-6.60}      & -3.00 / -3.70 / -1.05 / \textbf{-7.75}      \\ \hline
\end{tabular}
\begin{flushleft}
*The “- / - / - / -" indicates the missing, extra, reversed edges, and SHD obtained by the corresponding method in the row using the corresponding number of prior errors in the column.

\end{flushleft}
\label{table4}
\end{table*}
In this experiment, We compare various classic methods. Note that we focus more on the resilience of different methods to prior errors rather than the final SHD. This is because different methods have different basic performance, and it is unfair to directly compare SHD of different methods. Specifically, we calculate the changes in the metrics after adding priors, and the results are presented in Table \ref{table4}.

Some modifications are made to these methods to make them accept priors. For HC, all priors are integrated into a DAG as the search starting point. Meanwhile, during the search process, the deletion and reversal of edges corresponding to priors are prohibited. 
The PC algorithm first continuously removes non-existent edges from a completely undirected graph and then conducts orientation process for remaining edges. For PC, we skip the detection of the edges declared by priors in the deletion step, and orient edges as indicated by the prior before the orientation step begins. MMHC can be viewed as a combination of PC and HC. Thus, the modification of MMHC can be carried out in the same way as the modification of the PC and the HC, respectively. Finally, CaMML supports prior input and requires no modification.

The experimental results in Table \ref{table4} show that the SHD of the test cases of almost all other methods increases due to the impact of prior errors, while HC-QC shows the least increase. A particularly interesting point is that even when the set of priors contains only prior errors, HC-QC can yield smaller SHD results than those without priors in some specific cases, which may be surprising. The superior SHD results are mainly attributable to HC-QC's capability to modify prior errors to correct priors. The prior errors detected by HC-QC are either reversed or deleted according to a certain strategy. If prior errors are deemed to be removed, this is equivalent to reducing the number of prior errors, thus bringing the result closer to the state without adding prior errors. If the detected prior error is a reversed direct prior and is judged to be reversed, then it is equivalent to obtaining a correct prior. Similarly, if the prior error to be reversed is a reversed indirect prior, it is equivalent to adding an indirect prior for CSL. This action is also advantageous based on the property of indirect priors. Therefore, the proposed method may perform better even when the prior error comprises a large proportion of the priors set.

In some special cases, prior errors are less harmful, such as MMHC's results on  INSURANCE. The reason pertains to the reversed direct prior, which, as a prior, enforces the establishment of relationships between nodes so that some edges not found are counted as reverse edges. This reason applies to all methods. For instance, it can be observed that for PC or MMHC, the number of missing edges may decrease after adding prior errors. However, rather than expecting this phenomenon to occur, it is better to use QC-based method that is more resilient to prior errors.

\subsubsection{Comparison with cutting-edge methods}
\label{section_comparison_cutting-edge_Methods}

In this experiment, we tested NOTEARS-QC and various novel methods, comprehensively comparing the performance under different settings and configurations, focusing on the performance measured by F1 score and SHD. The results are shown in Table \ref{gradient_compare}.
The table is organized into several sections based on the number of nodes (30 and 40), different edges ratios to nodes  (ER=1 and ER=2), and different datasize ratios to nodes (DR=10 and DR=20). For each setting, results are provided for F1 score and SHD at three correct prior proportion (10\%, 20\% and 30\%) fot two prior error ratios (0.5, 1) respect to prior proportion.

The results show that performance of F1 score and SHD is sensitive to the prior error ratio, with higher ratios generally leading to poorer performance. This indicates that even for the CSL methods proposed in recent years, accurate prior is important to pay attention to. Although the NOTEARS is more error-tolerant compared to other methods, the application of Quasi-circle correction  to the NOTEARS still demonstrates significant improvements in performance, as evidenced by higher F1 scores and lower SHD values across various correct prior proportion and corresponding prior error ratios. This illustrates that the proposed strategy can improve the ability of the base method against prior errors over a wide range of experimental parameter spaces, justifying the critical role of proposed strategy in improving the accuracy of CSL from data.

\begin{table*}[]
\centering
\caption{Performance Comparison of Gradient-Based Causal Structure Learning Methods with Quasi-Circles-Based Prior Correction}
\begin{tabular}{cccc|cccc|cccc|cccc}
\multicolumn{4}{c|}{Correct Prior Proportion}                                                                                                                                                                & \multicolumn{4}{c|}{10\%}                                                           & \multicolumn{4}{c|}{20\%}                                                         & \multicolumn{4}{c}{30\%}                                                           \\ \hline
\multicolumn{4}{c|}{Prior Error Ratio}                                                                                                                                                               & \multicolumn{2}{c|}{0.5}                           & \multicolumn{2}{c|}{1}        & \multicolumn{2}{c|}{0.5}                          & \multicolumn{2}{c|}{1}       & \multicolumn{2}{c|}{0.5}                          & \multicolumn{2}{c}{1}         \\ \hline
\multicolumn{4}{c|}{Metric}                                                                                                                                                                          & F1             & \multicolumn{1}{c|}{SHD}          & F1             & SHD          & F1            & \multicolumn{1}{c|}{SHD}          & F1            & SHD          & F1            & \multicolumn{1}{c|}{SHD}          & F1            & SHD           \\ \hline
\multicolumn{1}{c|}{\multirow{24}{*}{\begin{tabular}[c]{@{}c@{}}Node\\      =30\end{tabular}}} & \multicolumn{1}{c|}{\multirow{12}{*}{ER=1}} & \multicolumn{1}{c|}{\multirow{6}{*}{DR=10}} & GranDAG & 21.5           & \multicolumn{1}{c|}{37.2}         & 20.5           & 38.3         & 33.7          & \multicolumn{1}{c|}{35.0}         & 29.4          & 38.3         & 37.7          & \multicolumn{1}{c|}{32.3}         & 34.5          & 36.7          \\
\multicolumn{1}{c|}{}                                                                          & \multicolumn{1}{c|}{}                       & \multicolumn{1}{c|}{}                       & GAE     & 44.3           & \multicolumn{1}{c|}{25.0}         & 42.5           & 26.5         & 48.5          & \multicolumn{1}{c|}{22.2}         & 46.8          & 23.0         & 44.5          & \multicolumn{1}{c|}{23.7}         & 42.9          & 24.7          \\
\multicolumn{1}{c|}{}                                                                          & \multicolumn{1}{c|}{}                       & \multicolumn{1}{c|}{}                       & DAG-GNN & 61.5           & \multicolumn{1}{c|}{27.5}         & 58.5           & 29.5         & 50.1          & \multicolumn{1}{c|}{38.2}         & 45.4          & 41.3         & 43.2          & \multicolumn{1}{c|}{42.8}         & 34.8          & 50.0          \\
\multicolumn{1}{c|}{}                                                                          & \multicolumn{1}{c|}{}                       & \multicolumn{1}{c|}{}                       & GOLEM   & 82.1           & \multicolumn{1}{c|}{8.3}          & 77.7           & 10.3         & 76.9          & \multicolumn{1}{c|}{10.7}         & 69.2          & 13.3         & 72.0          & \multicolumn{1}{c|}{12.7}         & 56.1          & 17.5          \\
\multicolumn{1}{c|}{}                                                                          & \multicolumn{1}{c|}{}                       & \multicolumn{1}{c|}{}                       & NOTEARS & 91.6           & \multicolumn{1}{c|}{4.3}          & 87.0           & 6.3          & 84.4          & \multicolumn{1}{c|}{7.0}          & 83.0          & 6.7          & 85.3          & \multicolumn{1}{c|}{6.3}          & 76.0          & 9.7           \\
\multicolumn{1}{c|}{}                                                                          & \multicolumn{1}{c|}{}                       & \multicolumn{1}{c|}{}                       & +QC     & \textbf{100.0} & \multicolumn{1}{c|}{\textbf{0.0}} & \textbf{100.0} & \textbf{0.0} & \textbf{95.6} & \multicolumn{1}{c|}{\textbf{1.5}} & \textbf{87.0} & \textbf{4.7} & \textbf{87.8} & \multicolumn{1}{c|}{\textbf{5.0}} & \textbf{80.4} & \textbf{7.7}  \\ \cline{3-16} 
\multicolumn{1}{c|}{}                                                                          & \multicolumn{1}{c|}{}                       & \multicolumn{1}{c|}{\multirow{6}{*}{DR=20}} & GranDAG & 22.3           & \multicolumn{1}{c|}{33.2}         & 21.8           & 33.0         & 34.5          & \multicolumn{1}{c|}{34.0}         & 30.4          & 36.5         & 41.5          & \multicolumn{1}{c|}{25.5}         & 38.8          & 29.2          \\
\multicolumn{1}{c|}{}                                                                          & \multicolumn{1}{c|}{}                       & \multicolumn{1}{c|}{}                       & GAE     & 39.5           & \multicolumn{1}{c|}{26.5}         & 39.7           & 26.3         & 45.6          & \multicolumn{1}{c|}{22.7}         & 43.4          & 24.5         & 42.1          & \multicolumn{1}{c|}{24.0}         & 38.9          & 25.7          \\
\multicolumn{1}{c|}{}                                                                          & \multicolumn{1}{c|}{}                       & \multicolumn{1}{c|}{}                       & DAG-GNN & 57.4           & \multicolumn{1}{c|}{25.0}         & 54.7           & 27.0         & 48.7          & \multicolumn{1}{c|}{32.7}         & 41.6          & 38.3         & 43.2          & \multicolumn{1}{c|}{36.3}         & 35.6          & 42.3          \\
\multicolumn{1}{c|}{}                                                                          & \multicolumn{1}{c|}{}                       & \multicolumn{1}{c|}{}                       & GOLEM   & 82.5           & \multicolumn{1}{c|}{8.0}          & 78.1           & 10.0         & 77.9          & \multicolumn{1}{c|}{10.0}         & 71.7          & 12.3         & 74.0          & \multicolumn{1}{c|}{11.8}         & 55.3          & 18.0          \\
\multicolumn{1}{c|}{}                                                                          & \multicolumn{1}{c|}{}                       & \multicolumn{1}{c|}{}                       & NOTEARS & 92.3           & \multicolumn{1}{c|}{3.8}          & 87.7           & 5.8          & 85.3          & \multicolumn{1}{c|}{6.3}          & 82.5          & 7.0          & 85.5          & \multicolumn{1}{c|}{6.2}          & 75.0          & 10.2          \\
\multicolumn{1}{c|}{}                                                                          & \multicolumn{1}{c|}{}                       & \multicolumn{1}{c|}{}                       & +QC     & \textbf{100.0} & \multicolumn{1}{c|}{\textbf{0.0}} & \textbf{100.0} & \textbf{0.0} & \textbf{96.7} & \multicolumn{1}{c|}{\textbf{1.0}} & \textbf{87.8} & \textbf{4.3} & \textbf{91.0} & \multicolumn{1}{c|}{\textbf{3.3}} & \textbf{79.7} & \textbf{8.0}  \\ \cline{2-16} 
\multicolumn{1}{c|}{}                                                                          & \multicolumn{1}{c|}{\multirow{12}{*}{ER=2}} & \multicolumn{1}{c|}{\multirow{6}{*}{DR=10}} & GranDAG & 22.9           & \multicolumn{1}{c|}{64.2}         & 20.3           & 68.5         & 27.0          & \multicolumn{1}{c|}{68.2}         & 24.3          & 66.5         & 32.7          & \multicolumn{1}{c|}{67.0}         & 29.9          & 63.2          \\
\multicolumn{1}{c|}{}                                                                          & \multicolumn{1}{c|}{}                       & \multicolumn{1}{c|}{}                       & GAE     & 64.9           & \multicolumn{1}{c|}{35.5}         & 37.6           & 60.2         & 17.1          & \multicolumn{1}{c|}{57.8}         & 12.5          & 58.8         & 16.5          & \multicolumn{1}{c|}{63.5}         & 12.8          & 65.3          \\
\multicolumn{1}{c|}{}                                                                          & \multicolumn{1}{c|}{}                       & \multicolumn{1}{c|}{}                       & DAG-GNN & 50.4           & \multicolumn{1}{c|}{64.8}         & 46.2           & 68.5         & 47.7          & \multicolumn{1}{c|}{66.2}         & 40.3          & 75.8         & 49.3          & \multicolumn{1}{c|}{64.3}         & 35.6          & 95.2          \\
\multicolumn{1}{c|}{}                                                                          & \multicolumn{1}{c|}{}                       & \multicolumn{1}{c|}{}                       & GOLEM   & 83.6           & \multicolumn{1}{c|}{16.3}         & 70.5           & 29.3         & 72.0          & \multicolumn{1}{c|}{26.7}         & 59.8          & 33.7         & 63.7          & \multicolumn{1}{c|}{31.8}         & 46.3          & 45.0          \\
\multicolumn{1}{c|}{}                                                                          & \multicolumn{1}{c|}{}                       & \multicolumn{1}{c|}{}                       & NOTEARS & 89.9           & \multicolumn{1}{c|}{10.0}         & 86.4           & 11.8         & 84.7          & \multicolumn{1}{c|}{13.0}         & 80.4          & 17.2         & 83.6          & \multicolumn{1}{c|}{14.2}         & 72.0          & 26.3          \\
\multicolumn{1}{c|}{}                                                                          & \multicolumn{1}{c|}{}                       & \multicolumn{1}{c|}{}                       & +QC     & \textbf{97.6}  & \multicolumn{1}{c|}{\textbf{2.5}} & \textbf{95.9}  & \textbf{3.7} & \textbf{93.2} & \multicolumn{1}{c|}{\textbf{6.2}} & \textbf{90.9} & \textbf{7.8} & \textbf{95.4} & \multicolumn{1}{c|}{\textbf{4.7}} & \textbf{92.0} & \textbf{7.8}  \\ \cline{3-16} 
\multicolumn{1}{c|}{}                                                                          & \multicolumn{1}{c|}{}                       & \multicolumn{1}{c|}{\multirow{6}{*}{DR=20}} & GranDAG & 24.2           & \multicolumn{1}{c|}{68.0}         & 17.4           & 73.5         & 26.8          & \multicolumn{1}{c|}{63.3}         & 24.6          & 61.7         & 33.9          & \multicolumn{1}{c|}{65.2}         & 30.8          & 60.3          \\
\multicolumn{1}{c|}{}                                                                          & \multicolumn{1}{c|}{}                       & \multicolumn{1}{c|}{}                       & GAE     & 67.4           & \multicolumn{1}{c|}{33.5}         & 33.4           & 61.0         & 20.4          & \multicolumn{1}{c|}{54.7}         & 17.6          & 58.0         & 38.7          & \multicolumn{1}{c|}{50.7}         & 35.3          & 58.3          \\
\multicolumn{1}{c|}{}                                                                          & \multicolumn{1}{c|}{}                       & \multicolumn{1}{c|}{}                       & DAG-GNN & 43.3           & \multicolumn{1}{c|}{61.8}         & 37.9           & 67.0         & 40.7          & \multicolumn{1}{c|}{65.7}         & 38.6          & 67.2         & 50.2          & \multicolumn{1}{c|}{49.8}         & 30.8          & 82.8          \\
\multicolumn{1}{c|}{}                                                                          & \multicolumn{1}{c|}{}                       & \multicolumn{1}{c|}{}                       & GOLEM   & 85.5           & \multicolumn{1}{c|}{14.7}         & 71.1           & 28.5         & 72.0          & \multicolumn{1}{c|}{26.7}         & 61.9          & 31.3         & 67.5          & \multicolumn{1}{c|}{28.8}         & 46.6          & 45.0          \\
\multicolumn{1}{c|}{}                                                                          & \multicolumn{1}{c|}{}                       & \multicolumn{1}{c|}{}                       & NOTEARS & 92.5           & \multicolumn{1}{c|}{7.0}          & 86.4           & 12.0         & 85.0          & \multicolumn{1}{c|}{12.7}         & 81.2          & 16.0         & 84.4          & \multicolumn{1}{c|}{13.3}         & 73.1          & 25.2          \\
\multicolumn{1}{c|}{}                                                                          & \multicolumn{1}{c|}{}                       & \multicolumn{1}{c|}{}                       & +QC     & \textbf{98.4}  & \multicolumn{1}{c|}{\textbf{1.7}} & \textbf{98.1}  & \textbf{1.3} & \textbf{97.6} & \multicolumn{1}{c|}{\textbf{1.8}} & \textbf{92.2} & \textbf{7.0} & \textbf{99.7} & \multicolumn{1}{c|}{\textbf{0.3}} & \textbf{93.4} & \textbf{5.8}  \\ \hline
\multicolumn{1}{c|}{\multirow{24}{*}{\begin{tabular}[c]{@{}c@{}}Node\\      =40\end{tabular}}} & \multicolumn{1}{c|}{\multirow{12}{*}{ER=1}} & \multicolumn{1}{c|}{\multirow{6}{*}{DR=10}} & GranDAG & 30.6           & \multicolumn{1}{c|}{41.5}         & 26.4           & 47.5         & 27.8          & \multicolumn{1}{c|}{47.3}         & 20.8          & 44.2         & 32.6          & \multicolumn{1}{c|}{37.8}         & 29.9          & 42.0          \\
\multicolumn{1}{c|}{}                                                                          & \multicolumn{1}{c|}{}                       & \multicolumn{1}{c|}{}                       & GAE     & 48.4           & \multicolumn{1}{c|}{28.0}         & 37.5           & 32.0         & 27.4          & \multicolumn{1}{c|}{34.3}         & 26.4          & 34.7         & 40.4          & \multicolumn{1}{c|}{30.8}         & 31.0          & 34.5          \\
\multicolumn{1}{c|}{}                                                                          & \multicolumn{1}{c|}{}                       & \multicolumn{1}{c|}{}                       & DAG-GNN & 57.6           & \multicolumn{1}{c|}{32.7}         & 44.2           & 46.8         & 39.0          & \multicolumn{1}{c|}{53.3}         & 34.7          & 57.3         & 42.3          & \multicolumn{1}{c|}{47.3}         & 31.4          & 62.0          \\
\multicolumn{1}{c|}{}                                                                          & \multicolumn{1}{c|}{}                       & \multicolumn{1}{c|}{}                       & GOLEM   & 91.3           & \multicolumn{1}{c|}{6.0}          & 79.2           & 15.3         & 79.3          & \multicolumn{1}{c|}{14.5}         & 66.8          & 21.8         & 75.4          & \multicolumn{1}{c|}{14.3}         & 50.4          & 31.8          \\
\multicolumn{1}{c|}{}                                                                          & \multicolumn{1}{c|}{}                       & \multicolumn{1}{c|}{}                       & NOTEARS & 92.3           & \multicolumn{1}{c|}{5.3}          & 79.2           & 14.3         & 79.2          & \multicolumn{1}{c|}{14.3}         & 70.2          & 18.7         & 77.6          & \multicolumn{1}{c|}{13.5}         & 64.0          & 22.8          \\
\multicolumn{1}{c|}{}                                                                          & \multicolumn{1}{c|}{}                       & \multicolumn{1}{c|}{}                       & +QC     & \textbf{99.6}  & \multicolumn{1}{c|}{\textbf{0.3}} & \textbf{96.9}  & \textbf{1.5} & \textbf{96.8} & \multicolumn{1}{c|}{\textbf{1.5}} & \textbf{85.9} & \textbf{6.3} & \textbf{88.2} & \multicolumn{1}{c|}{\textbf{5.8}} & \textbf{83.0} & \textbf{8.5}  \\ \cline{3-16} 
\multicolumn{1}{c|}{}                                                                          & \multicolumn{1}{c|}{}                       & \multicolumn{1}{c|}{\multirow{6}{*}{DR=20}} & GranDAG & 32.6           & \multicolumn{1}{c|}{41.5}         & 29.9           & 44.7         & 29.4          & \multicolumn{1}{c|}{47.7}         & 21.1          & 42.5         & 33.4          & \multicolumn{1}{c|}{43.8}         & 29.7          & 45.2          \\
\multicolumn{1}{c|}{}                                                                          & \multicolumn{1}{c|}{}                       & \multicolumn{1}{c|}{}                       & GAE     & 31.8           & \multicolumn{1}{c|}{33.0}         & 25.5           & 34.3         & 23.8          & \multicolumn{1}{c|}{35.0}         & 31.8          & 34.3         & 48.4          & \multicolumn{1}{c|}{29.5}         & 42.1          & 30.8          \\
\multicolumn{1}{c|}{}                                                                          & \multicolumn{1}{c|}{}                       & \multicolumn{1}{c|}{}                       & DAG-GNN & 40.9           & \multicolumn{1}{c|}{36.3}         & 36.0           & 42.3         & 36.9          & \multicolumn{1}{c|}{40.2}         & 33.0          & 43.2         & 40.2          & \multicolumn{1}{c|}{37.7}         & 33.5          & 44.8          \\
\multicolumn{1}{c|}{}                                                                          & \multicolumn{1}{c|}{}                       & \multicolumn{1}{c|}{}                       & GOLEM   & 91.9           & \multicolumn{1}{c|}{5.7}          & 79.7           & 14.8         & 78.8          & \multicolumn{1}{c|}{15.0}         & 66.7          & 22.3         & 75.0          & \multicolumn{1}{c|}{14.7}         & 51.4          & 31.7          \\
\multicolumn{1}{c|}{}                                                                          & \multicolumn{1}{c|}{}                       & \multicolumn{1}{c|}{}                       & NOTEARS & 92.1           & \multicolumn{1}{c|}{5.5}          & 79.0           & 14.7         & 79.2          & \multicolumn{1}{c|}{14.3}         & 70.2          & 18.7         & 81.8          & \multicolumn{1}{c|}{10.5}         & 67.3          & 20.5          \\
\multicolumn{1}{c|}{}                                                                          & \multicolumn{1}{c|}{}                       & \multicolumn{1}{c|}{}                       & +QC     & \textbf{99.8}  & \multicolumn{1}{c|}{\textbf{0.2}} & \textbf{96.3}  & \textbf{1.8} & \textbf{97.1} & \multicolumn{1}{c|}{\textbf{1.3}} & \textbf{86.9} & \textbf{5.7} & \textbf{90.0} & \multicolumn{1}{c|}{\textbf{4.7}} & \textbf{86.0} & \textbf{6.8}  \\ \cline{2-16} 
\multicolumn{1}{c|}{}                                                                          & \multicolumn{1}{c|}{\multirow{12}{*}{ER=2}} & \multicolumn{1}{c|}{\multirow{6}{*}{DR=10}} & GranDAG & 28.5           & \multicolumn{1}{c|}{93.0}         & 28.5           & 96.7         & 34.1          & \multicolumn{1}{c|}{89.3}         & 31.1          & 78.2         & 43.0          & \multicolumn{1}{c|}{68.2}         & 38.1          & 72.0          \\
\multicolumn{1}{c|}{}                                                                          & \multicolumn{1}{c|}{}                       & \multicolumn{1}{c|}{}                       & GAE     & 51.2           & \multicolumn{1}{c|}{59.7}         & 24.6           & 75.0         & 56.9          & \multicolumn{1}{c|}{52.0}         & 43.2          & 63.7         & 51.0          & \multicolumn{1}{c|}{56.7}         & 40.6          & 71.3          \\
\multicolumn{1}{c|}{}                                                                          & \multicolumn{1}{c|}{}                       & \multicolumn{1}{c|}{}                       & DAG-GNN & 57.9           & \multicolumn{1}{c|}{80.2}         & 51.4           & 99.0         & 49.2          & \multicolumn{1}{c|}{112.7}        & 36.4          & 150.5        & 43.9          & \multicolumn{1}{c|}{127.5}        & 33.9          & 148.7         \\
\multicolumn{1}{c|}{}                                                                          & \multicolumn{1}{c|}{}                       & \multicolumn{1}{c|}{}                       & GOLEM   & 82.4           & \multicolumn{1}{c|}{25.8}         & 63.3           & 48.0         & 75.8          & \multicolumn{1}{c|}{32.2}         & 52.1          & 66.7         & 68.0          & \multicolumn{1}{c|}{41.5}         & 51.0          & 53.2          \\
\multicolumn{1}{c|}{}                                                                          & \multicolumn{1}{c|}{}                       & \multicolumn{1}{c|}{}                       & NOTEARS & 85.2           & \multicolumn{1}{c|}{22.0}         & 82.1           & 24.7         & 85.2          & \multicolumn{1}{c|}{18.8}         & 68.2          & 50.5         & 78.7          & \multicolumn{1}{c|}{30.3}         & 68.3          & 49.2          \\
\multicolumn{1}{c|}{}                                                                          & \multicolumn{1}{c|}{}                       & \multicolumn{1}{c|}{}                       & +QC     & \textbf{98.2}  & \multicolumn{1}{c|}{\textbf{2.5}} & \textbf{97.7}  & \textbf{3.3} & \textbf{98.1} & \multicolumn{1}{c|}{\textbf{2.5}} & \textbf{93.3} & \textbf{9.7} & \textbf{94.5} & \multicolumn{1}{c|}{\textbf{6.5}} & \textbf{91.5} & \textbf{11.0} \\ \cline{3-16} 
\multicolumn{1}{c|}{}                                                                          & \multicolumn{1}{c|}{}                       & \multicolumn{1}{c|}{\multirow{6}{*}{DR=20}} & GranDAG & 33.9           & \multicolumn{1}{c|}{94.0}         & 30.9           & 101.0        & 34.2          & \multicolumn{1}{c|}{85.7}         & 30.7          & 77.7         & 41.6          & \multicolumn{1}{c|}{70.5}         & 39.5          & 70.3          \\
\multicolumn{1}{c|}{}                                                                          & \multicolumn{1}{c|}{}                       & \multicolumn{1}{c|}{}                       & GAE     & 52.8           & \multicolumn{1}{c|}{60.7}         & 13.8           & 92.5         & 54.7          & \multicolumn{1}{c|}{54.7}         & 45.6          & 61.3         & 51.0          & \multicolumn{1}{c|}{58.0}         & 34.6          & 74.7          \\
\multicolumn{1}{c|}{}                                                                          & \multicolumn{1}{c|}{}                       & \multicolumn{1}{c|}{}                       & DAG-GNN & 60.0           & \multicolumn{1}{c|}{63.2}         & 50.9           & 82.8         & 45.3          & \multicolumn{1}{c|}{101.0}        & 33.9          & 137.3        & 42.9          & \multicolumn{1}{c|}{109.0}        & 32.0          & 121.2         \\
\multicolumn{1}{c|}{}                                                                          & \multicolumn{1}{c|}{}                       & \multicolumn{1}{c|}{}                       & GOLEM   & 83.7           & \multicolumn{1}{c|}{24.0}         & 60.1           & 53.2         & 74.1          & \multicolumn{1}{c|}{34.3}         & 53.4          & 63.8         & 67.8          & \multicolumn{1}{c|}{41.7}         & 48.9          & 59.3          \\
\multicolumn{1}{c|}{}                                                                          & \multicolumn{1}{c|}{}                       & \multicolumn{1}{c|}{}                       & NOTEARS & 88.2           & \multicolumn{1}{c|}{16.3}         & 83.3           & 22.7         & 83.0          & \multicolumn{1}{c|}{22.2}         & 66.8          & 53.7         & 79.1          & \multicolumn{1}{c|}{30.3}         & 67.9          & 50.3          \\
\multicolumn{1}{c|}{}                                                                          & \multicolumn{1}{c|}{}                       & \multicolumn{1}{c|}{}                       & +QC     & \textbf{98.5}  & \multicolumn{1}{c|}{\textbf{1.7}} & \textbf{99.0}  & \textbf{1.3} & \textbf{98.4} & \multicolumn{1}{c|}{\textbf{2.0}} & \textbf{93.9} & \textbf{8.8} & \textbf{94.7} & \multicolumn{1}{c|}{\textbf{6.7}} & \textbf{88.9} & \textbf{14.8}
\end{tabular}
\begin{flushleft}
*ER represents the ratio of the number of edges to nodes, and DR represents the ratio of datasize to the number of nodes. ``+QC" indicates NOTEARS-QC which utilizes proposed prior correction strategy.
\end{flushleft}
\label{gradient_compare}
\end{table*}

\subsection{The Resilience Test of Different Methods to Prior Errors.}

The results of the above experiments have demonstrated that a prior set containing numerous errors may be inferior to no prior, as the former may lead to worse results. Therefore, to enhance the accuracy of causal structure learning, the correct prior should occupy a sufficient proportion in the priors set. 
The proposed quasi-circle-based correction strategy reduces this requirement and can tolerate lower quality priors (prior quality is defined as the proportion of correct priors in the prior set). In this section, we experimentally explore the lower bounds of the prior quality of different methods. Experiments commence with a set of priors containing only prior errors. By continuously adding correct priors to this prior set until the obtained SHD can return to the level without priors, the ratio of correct priors to the total number of priors is obtained as the lower bound of the prior quality that the corresponding method can tolerate. We conducted experiments using HC-QC and various classic methods. Each method was conducted a total of 10 times, and the experimental results are shown in Table \ref{table5}.

\begin{table}[htbp]
\centering
\setlength{\tabcolsep}{4pt}
\caption{The Lower Bounds on the Required Prior Quality of Different Methods}
\begin{tabular}{@{}c@{}|@{}c@{}||c@{}c@{}c@{}c@{}||c@{}c@{}c@{}c@{}}
\hline
\multirow{2}{*}{Network}   & \multirow{2}{*}{\begin{tabular}[c]{@{}c@{}}
     Priors  \\
     Errors
\end{tabular} } & \multicolumn{4}{c||}{Number of Correct Priors} & \multicolumn{4}{c}{Prior Quality}   \\ \cline{3-10} 
                           &                                  & PC       & MMHC     & HC       & HC-QC    & PC     & MMHC   & HC     & HC-QC \\ \hline \hline
\multirow{5}{*}{Child}     & 1                                & 3.2      & 3.9      & 2.6      & \textbf{2.0}         & 76.2 & 79.6 & 72.2 & \textbf{66.1}   \\
                           & 2                                & 5.9      & 5.4      & 3.7      & \textbf{2.6}         & 74.7 & 73.0 & 64.9 & \textbf{56.5}   \\
                           & 3                                & 8.3      & 6.1      & 5.1      & \textbf{3.0}         & 73.3 & 66.9 & 63.0 & \textbf{50.0}   \\
                           & 4                                & 11.0     & 7.0      & 5.6      & \textbf{3.7}         & 73.3 & 63.5 & 58.3 & \textbf{47.7}   \\
                           & 5                                & 12.5     & 10.1     & 11.6     & \textbf{6.9}         & 71.3 & 66.8 & 69.8 & \textbf{57.8}   \\ \hline\hline
\multirow{5}{*}{Alarm}     & 1                                & 8.4      & 9.4      & \textbf{4.1}      & 5.9         & 89.3 & 90.4 & \textbf{80.2} & 85.5   \\
                           & 2                                & 14.2     & 10.8     & 5.9      & \textbf{5.2}         & 87.6 & 84.4 & 74.5 & \textbf{72.0}   \\
                           & 3                                & 20.8     & 11.9     & 8.7      & \textbf{5.1}         & 87.4 & 79.8 & 74.4 & \textbf{63.0}   \\
                           & 4                                & 25.5     & 14.9     & 10.2     & \textbf{5.6}         & 86.4 & 78.8 & 71.7 & \textbf{58.3}   \\
                           & 5                                & 29.2     & 17.1     & 13.0     & \textbf{7.9}         & 85.4 & 77.3 & 72.1 & \textbf{61.2}   \\ \hline\hline
\multirow{5}{*}{Insurance} & 1                                & 2.4      & 1.0      & 0.6      & \textbf{0.5}         & 70.6 & 48.7 & 37.5 & \textbf{33.3}   \\
                           & 2                                & 4.1      & 1.4      & \textbf{0.6}      & 0.7         & 66.9 & 40.3 & \textbf{23.1} & 24.5   \\
                           & 3                                & 4.9      & 1.7      & 1.1      & \textbf{0.8}         & 62.0 & 36.2 & 26.8 & \textbf{21.1}   \\
                           & 4                                & 6.2      & 2.4      & 1.3      & \textbf{0.7}         & 60.6 & 37.0 & 23.8 & \textbf{14.9}   \\
                           & 5                                & 7.9      & 2.5      & 1.6      & \textbf{0.8}         & 61.1 & 33.3 & 24.2 & \textbf{13.0}   \\ \hline
\end{tabular}
\label{table5}
\end{table}

Experimental results show that the larger the number of prior errors, the more correct priors are needed, and the growth rate of required correct priors may reach several times that of prior errors. From the perspective of prior quality, this means that the application of priors in real scenarios is challenging to achieve without high-quality priors as conditions. Although the prior quality requirement decreases slightly (about 10\%) as the set of priors becomes larger, it is still necessary to have a majority of correct priors. In contrast, HC-QC can tolerate worse prior quality, which creates a certain fault-tolerant possibility for the upstream prior acquisition method. Some networks may be sensitive to initial input, where correct initial priors can substantially improve results, preventing prior errors or score guidance from directing the DAG search in the wrong direction. The INSURANCE network might be an example of this, as it turns out that INSURANCE requires fewer correct priors than other networks. For such networks, HC-QC may achieve superior results. This is due to the property of HC-QC to convert incorrect direct priors or incorrect indirect priors into beneficial priors, thus requiring fewer correct priors.

\subsection{Future Work}
This paper employs quasi-circles to detect edge priors. However, there are some more complex priors used by the field of causal structure learning, such as structural or conditional priors. Naturally, errors in these priors may lead to a large number of structural errors. Consequently, future research is warranted to address the detection of these types of priors.

Furthermore, in many scenarios such as knowledge graphs and expert systems \cite{wang2021knowledge,wu2015knowledge,9885030}, priors may not be readily provided by domain experts, but require a certain extraction process \cite{wang2023distribution}. Unreliable priors may severely compromise the performance of causal structure learning methods. In such contexts, the method proposed in this paper could prove to be highly valuable, thereby constituting a significant future direction of this work.

\section{Conclusions}
\label{section_conclusions}

To address the issue of prior quality in causal structure learning, this paper investigates the theoretical properties of diverse prior errors. Subsequently, a post-hoc strategy is introduced, which is robust against edge-level prior errors and does not require supervision. This strategy is not confined to a specific causal learning process and can be readily integrated into existing causal structure learning methods to create a version resilient against unreliable priors.

Experiments have been conducted on various Bayesian network datasets, demonstrating the proposed strategy's capacity to resist order-reversed prior errors while preserving the correct prior. This performance surpasses existing methods capable of identifying prior errors. Further, through comparative experiments with a range of widely used methods, the proposed strategy's performance improvement in CSL is clearly illustrated. The findings reveal that the strategy proposed in this paper can tolerate a lower quality of priors compared to other methods. This tolerance can mitigate the stringent demands of prior acquisition and facilitate the practical application of causal theory in real-world scenarios.

\bibliographystyle{IEEEtran}

\footnotesize
\bibliography{ref}

\begin{thebibliography}{10}
\providecommand{\url}[1]{#1}
\csname url@samestyle\endcsname
\providecommand{\newblock}{\relax}
\providecommand{\bibinfo}[2]{#2}
\providecommand{\BIBentrySTDinterwordspacing}{\spaceskip=0pt\relax}
\providecommand{\BIBentryALTinterwordstretchfactor}{4}
\providecommand{\BIBentryALTinterwordspacing}{\spaceskip=\fontdimen2\font plus
\BIBentryALTinterwordstretchfactor\fontdimen3\font minus \fontdimen4\font\relax}
\providecommand{\BIBforeignlanguage}[2]{{%
\expandafter\ifx\csname l@#1\endcsname\relax
\typeout{** WARNING: IEEEtran.bst: No hyphenation pattern has been}%
\typeout{** loaded for the language `#1'. Using the pattern for}%
\typeout{** the default language instead.}%
\else
\language=\csname l@#1\endcsname
\fi
#2}}
\providecommand{\BIBdecl}{\relax}
\BIBdecl

\bibitem{zhang2021survey}
Y.~Zhang, P.~Ti{\v{n}}o, A.~Leonardis, and K.~Tang, ``A survey on neural network interpretability,'' \emph{IEEE Transactions on Emerging Topics in Computational Intelligence}, vol.~5, no.~5, pp. 726--742, 2021.

\bibitem{xin_wang}
X.~Wang, S.~Lyu, X.~Wu, T.~Wu, and H.~Chen, ``Generalization bounds for estimating causal effects of continuous treatments,'' in \emph{Advances in Neural Information Processing Systems}, vol.~35, 2022, pp. 8605--8617.

\bibitem{yu2019multi}
K.~Yu, L.~Liu, J.~Li, W.~Ding, and T.~D. Le, ``Multi-source causal feature selection,'' \emph{IEEE Transactions on Pattern Analysis and Machine Intelligence}, vol.~42, no.~9, pp. 2240--2256, 2019.

\bibitem{chu2023machine}
S.~Chu, A.~Jiang, L.~Chen, X.~Zhang, X.~Shen, W.~Zhou, S.~Ye, C.~Chen, S.~Zhang, L.~Zhang \emph{et~al.}, ``Machine learning algorithms for predicting the risk of fracture in patients with diabetes in china,'' \emph{Heliyon}, vol.~9, no.~7, 2023.

\bibitem{9756301}
H.~Zhang, L.~Xiao, X.~Cao, and H.~Foroosh, ``Multiple adverse weather conditions adaptation for object detection via causal intervention,'' \emph{IEEE Transactions on Pattern Analysis and Machine Intelligence}, pp. 1--1, 2022.

\bibitem{scanagatta2019survey}
M.~Scanagatta, A.~Salmer{\'o}n, and F.~Stella, ``A survey on {B}ayesian network structure learning from data,'' \emph{Progress in Artificial Intelligence}, vol.~8, pp. 425--439, 2019.

\bibitem{jiang2019joint}
B.~Jiang, X.~Wu, K.~Yu, and H.~Chen, ``Joint {S}emi-supervised feature selection and classification through {B}ayesian approach,'' in \emph{Proceedings of the AAAI Conference on Artificial Intelligence}, vol.~33, no.~01, 2019, pp. 3983--3990.

\bibitem{li2014sparse}
C.~Li and H.~Chen, ``Sparse {B}ayesian approach for feature selection,'' in \emph{2014 IEEE Symposium on Computational Intelligence in Big Data}.\hskip 1em plus 0.5em minus 0.4em\relax IEEE, 2014, pp. 1--7.

\bibitem{de2011efficient}
C.~P. De~Campos and Q.~Ji, ``Efficient structure learning of {Bayesian} networks using constraints,'' \emph{The Journal of Machine Learning Research}, vol.~12, pp. 663--689, 2011.

\bibitem{wang2023accurate}
X.~Wang, L.~Chen, T.~Ban, D.~Lyu, Y.~Guan, X.~Wu, X.~Zhou, and H.~Chen, ``Accurate label refinement from multiannotator of remote sensing data,'' \emph{IEEE Transactions on Geoscience and Remote Sensing}, vol.~61, pp. 1--13, 2023.

\bibitem{9815029}
T.~Ban, X.~Wang, L.~Chen, X.~Wu, Q.~Chen, and H.~Chen, ``Quality evaluation of triples in knowledge graph by incorporating internal with external consistency,'' \emph{IEEE Transactions on Neural Networks and Learning Systems}, pp. 1--13, 2022.

\bibitem{constantinou2021impact}
A.~C. Constantinou, Z.~Guo, and N.~K. Kitson, ``The impact of prior knowledge on causal structure learning,'' \emph{arXiv preprint arXiv:2102.00473}, 2021.

\bibitem{de2008improving}
C.~P. De~Campos and Q.~Ji, ``Improving {Bayesian} network parameter learning using constraints,'' in \emph{2008 19th International Conference on Pattern Recognition}.\hskip 1em plus 0.5em minus 0.4em\relax IEEE, 2008, pp. 1--4.

\bibitem{asvatourian2020integrating}
V.~Asvatourian, P.~Leray, S.~Michiels, and E.~Lanoy, ``Integrating expert’s knowledge constraint of time dependent exposures in structure learning for {Bayesian} networks,'' \emph{Artificial Intelligence in Medicine}, vol. 107, p. 101874, 2020.

\bibitem{9583890}
X.~Yang, H.~Zhang, and J.~Cai, ``Deconfounded image captioning: A causal retrospect,'' \emph{IEEE Transactions on Pattern Analysis and Machine Intelligence}, pp. 1--1, 2021.

\bibitem{9018179}
P.~N. Garner and S.~Tong, ``A {B}ayesian approach to recurrence in neural networks,'' \emph{IEEE Transactions on Pattern Analysis and Machine Intelligence}, vol.~43, no.~8, pp. 2527--2537, 2021.

\bibitem{jiang2017scalable}
B.~Jiang, H.~Chen, B.~Yuan, and X.~Yao, ``Scalable graph-based {S}emi-supervised learning through sparse {B}ayesian model,'' \emph{IEEE Transactions on Knowledge and Data Engineering}, vol.~29, no.~12, pp. 2758--2771, 2017.

\bibitem{long2023can}
S.~Long, T.~Schuster, A.~Pich{\'e}, S.~Research \emph{et~al.}, ``Can large language models build causal graphs?'' \emph{arXiv preprint arXiv:2303.05279}, 2023.

\bibitem{ban2023causal}
T.~Ban, L.~Chen, D.~Lyu, X.~Wang, and H.~Chen, ``Causal structure learning supervised by large language model,'' \emph{arXiv preprint arXiv:2311.11689}, 2023.

\bibitem{kiciman2023causal}
E.~K{\i}c{\i}man, R.~Ness, A.~Sharma, and C.~Tan, ``Causal reasoning and large language models: Opening a new frontier for causality,'' \emph{arXiv preprint arXiv:2305.00050}, 2023.

\bibitem{ban2023query}
T.~Ban, L.~Chen, X.~Wang, and H.~Chen, ``From query tools to causal architects: Harnessing large language models for advanced causal discovery from data,'' \emph{arXiv preprint arXiv:2306.16902}, 2023.

\bibitem{vowels2022d}
M.~J. Vowels, N.~C. Camgoz, and R.~Bowden, ``D’ya like dags? a survey on structure learning and causal discovery,'' \emph{ACM Computing Surveys}, vol.~55, no.~4, pp. 1--36, 2022.

\bibitem{kitson2023survey}
N.~K. Kitson, A.~C. Constantinou, Z.~Guo, Y.~Liu, and K.~Chobtham, ``A survey of bayesian network structure learning,'' \emph{Artificial Intelligence Review}, pp. 1--94, 2023.

\bibitem{o2006causal}
R.~T. O’Donnell, A.~E. Nicholson, B.~Han, K.~B. Korb, M.~J. Alam, and L.~R. Hope, ``Causal discovery with prior information,'' in \emph{AI 2006: Advances in Artificial Intelligence: 19th Australian Joint Conference on Artificial Intelligence, Hobart, Australia, December 4-8, 2006. Proceedings 19}.\hskip 1em plus 0.5em minus 0.4em\relax Springer, 2006, pp. 1162--1167.

\bibitem{tsagris2019bayesian}
M.~Tsagris, ``Bayesian network learning with the {PC} algorithm: {An} improved and correct variation,'' \emph{Applied Artificial Intelligence}, vol.~33, no.~2, pp. 101--123, 2019.

\bibitem{zheng2018dags}
X.~Zheng, B.~Aragam, P.~K. Ravikumar, and E.~P. Xing, ``Dags with no tears: Continuous optimization for structure learning,'' \emph{Advances in neural information processing systems}, vol.~31, 2018.

\bibitem{tsamardinos2006max}
I.~Tsamardinos, L.~E. Brown, and C.~F. Aliferis, ``The {Max-Min Hill-Climbing} {Bayesian} network structure learning algorithm,'' \emph{Machine Learning}, vol.~65, pp. 31--78, 2006.

\bibitem{castelo2000priors}
R.~Castelo and A.~Siebes, ``Priors on network structures. {Biasing} the search for {Bayesian} networks,'' \emph{International Journal of Approximate Reasoning}, vol.~24, no.~1, pp. 39--57, 2000.

\bibitem{amirkhani2016exploiting}
H.~Amirkhani, M.~Rahmati, P.~J. Lucas, and A.~Hommersom, ``Exploiting experts{'} knowledge for structure learning of {Bayesian} networks,'' \emph{IEEE Transactions on Pattern Analysis and Machine Intelligence}, vol.~39, no.~11, pp. 2154--2170, 2016.

\bibitem{borboudakis2014scoring}
G.~Borboudakis and I.~Tsamardinos, ``Scoring and searching over {Bayesian} networks with causal and associative priors,'' \emph{arXiv preprint arXiv:1408.2057}, 2014.

\bibitem{eggeling2019structure}
R.~Eggeling, J.~Viinikka, A.~Vuoksenmaa, and M.~Koivisto, ``On structure priors for learning {Bayesian} networks,'' in \emph{The 22nd International Conference on Artificial Intelligence and Statistics}.\hskip 1em plus 0.5em minus 0.4em\relax PMLR, 2019, pp. 1687--1695.

\bibitem{borboudakis2012incorporating}
G.~Borboudakis and I.~Tsamardinos, ``Incorporating causal prior knowledge as path-constraints in {Bayesian} networks and maximal ancestral graphs,'' \emph{arXiv preprint arXiv:1206.6390}, 2012.

\bibitem{de2007bayesian}
L.~M. de~Campos and J.~G. Castellano, ``Bayesian network learning algorithms using structural restrictions,'' \emph{International Journal of Approximate Reasoning}, vol.~45, no.~2, pp. 233--254, 2007.

\bibitem{de2009structure}
C.~P. De~Campos, Z.~Zeng, and Q.~Ji, ``Structure learning of {Bayesian} networks using constraints,'' in \emph{Proceedings of the 26th Annual International Conference on Machine Learning}, 2009, pp. 113--120.

\bibitem{chen2016learning}
E.~Y.-J. Chen, Y.~Shen, A.~Choi, and A.~Darwiche, ``Learning {Bayesian} networks with ancestral constraints,'' \emph{Advances in Neural Information Processing Systems}, vol.~29, 2016.

\bibitem{wang2021learning}
Z.~Wang, X.~Gao, Y.~Yang, X.~Tan, and D.~Chen, ``Learning {Bayesian} networks based on order graph with ancestral constraints,'' \emph{Knowledge-Based Systems}, vol. 211, p. 106515, 2021.

\bibitem{li2018bayesian}
A.~Li and P.~Beek, ``Bayesian network structure learning with side constraints,'' in \emph{International Conference on Probabilistic Graphical Models}.\hskip 1em plus 0.5em minus 0.4em\relax PMLR, 2018, pp. 225--236.

\bibitem{cano2011method}
A.~Cano, A.~R. Masegosa, and S.~Moral, ``A method for integrating expert knowledge when learning {B}ayesian networks from data,'' \emph{IEEE Transactions on Systems, Man, and Cybernetics, Part B (Cybernetics)}, vol.~41, no.~5, pp. 1382--1394, 2011.

\bibitem{7364252}
L.~Zhou, L.~Wang, L.~Liu, P.~Ogunbona, and D.~Shen, ``Learning discriminative {B}ayesian networks from high-dimensional continuous neuroimaging data,'' \emph{IEEE Transactions on Pattern Analysis and Machine Intelligence}, vol.~38, no.~11, pp. 2269--2283, 2016.

\bibitem{9079582}
Q.~Ye, A.~A. Amini, and Q.~Zhou, ``Optimizing regularized {C}holesky score for order-based learning of {B}ayesian networks,'' \emph{IEEE Transactions on Pattern Analysis and Machine Intelligence}, vol.~43, no.~10, pp. 3555--3572, 2021.

\bibitem{neath2012bayesian}
A.~A. Neath and J.~E. Cavanaugh, ``The {Bayesian} information criterion: {Background}, derivation, and applications,'' \emph{Wiley Interdisciplinary Reviews: Computational Statistics}, vol.~4, no.~2, pp. 199--203, 2012.

\bibitem{suzuki2017theoretical}
J.~Suzuki, ``A theoretical analysis of the {BDeu} scores in {Bayesian} network structure learning,'' \emph{Behaviormetrika}, vol.~44, pp. 97--116, 2017.

\bibitem{hansen2001model}
M.~H. Hansen and B.~Yu, ``Model selection and the principle of minimum description length,'' \emph{Journal of the American Statistical Association}, vol.~96, no. 454, pp. 746--774, 2001.

\bibitem{malinsky2019learning}
D.~Malinsky and P.~Spirtes, ``Learning the structure of a nonstationary vector autoregression,'' in \emph{The 22nd International Conference on Artificial Intelligence and Statistics}.\hskip 1em plus 0.5em minus 0.4em\relax PMLR, 2019, pp. 2986--2994.

\bibitem{gao2018parallel}
T.~Gao and D.~Wei, ``Parallel {Bayesian} network structure learning,'' in \emph{International Conference on Machine Learning}.\hskip 1em plus 0.5em minus 0.4em\relax PMLR, 2018, pp. 1685--1694.

\bibitem{chickering2002optimal}
D.~M. Chickering, ``Optimal structure identification with greedy search,'' \emph{Journal of Machine Learning Research}, vol.~3, no. Nov, pp. 507--554, 2002.

\bibitem{lachapelle2019gradient}
S.~Lachapelle, P.~Brouillard, T.~Deleu, and S.~Lacoste-Julien, ``Gradient-based neural {DAG} learning,'' \emph{arXiv preprint arXiv:1906.02226}, 2019.

\bibitem{ng2019graph}
I.~Ng, S.~Zhu, Z.~Chen, and Z.~Fang, ``A graph autoencoder approach to causal structure learning,'' \emph{arXiv preprint arXiv:1911.07420}, 2019.

\bibitem{yu2019dag}
Y.~Yu, J.~Chen, T.~Gao, and M.~Yu, ``{DAG-GNN}: {DAG} structure learning with graph neural networks,'' in \emph{International Conference on Machine Learning}.\hskip 1em plus 0.5em minus 0.4em\relax PMLR, 2019, pp. 7154--7163.

\bibitem{ng2020role}
I.~Ng, A.~Ghassami, and K.~Zhang, ``On the role of sparsity and {DAG} constraints for learning linear {DAGs},'' \emph{Advances in Neural Information Processing Systems}, vol.~33, pp. 17\,943--17\,954, 2020.

\bibitem{sachs2005causal}
K.~Sachs, O.~Perez, D.~Pe'er, D.~A. Lauffenburger, and G.~P. Nolan, ``Causal protein-signaling networks derived from multiparameter single-cell data,'' \emph{Science}, vol. 308, no. 5721, pp. 523--529, 2005.

\bibitem{schafer2005shrinkage}
J.~Sch{\"a}fer and K.~Strimmer, ``A shrinkage approach to large-scale covariance matrix estimation and implications for functional genomics,'' \emph{Statistical applications in genetics and molecular biology}, vol.~4, no.~1, 2005.

\bibitem{scutari2014multiple}
M.~Scutari, P.~Howell, D.~J. Balding, and I.~Mackay, ``Multiple quantitative trait analysis using {B}ayesian networks,'' \emph{Genetics}, vol. 198, no.~1, pp. 129--137, 2014.

\bibitem{martinez2020computational}
C.~Mart{\'\i}nez-Mart{\'\i}nez, J.~Mendez-Bermudez, J.~M. Rodr{\'\i}guez, and J.~M. Sigarreta, ``Computational and analytical studies of the {R}andi{\'c} index in {E}rd{\"o}s--{R}{\'e}nyi models,'' \emph{Applied Mathematics and Computation}, vol. 377, p. 125137, 2020.

\bibitem{wang2021knowledge}
X.~Wang, L.~Chen, T.~Ban, M.~Usman, Y.~Guan, S.~Liu, T.~Wu, and H.~Chen, ``Knowledge graph quality control: A survey,'' \emph{Fundamental Research}, vol.~1, no.~5, pp. 607--626, 2021.

\bibitem{wu2015knowledge}
X.~Wu, H.~Chen, G.~Wu, J.~Liu, Q.~Zheng, X.~He, A.~Zhou, Z.-Q. Zhao, B.~Wei, M.~Gao \emph{et~al.}, ``Knowledge engineering with big data,'' \emph{IEEE Intelligent Systems}, vol.~30, no.~5, pp. 46--55, 2015.

\bibitem{9885030}
X.~Wang, T.~Ban, L.~Chen, X.~Wu, D.~Lyu, and H.~Chen, ``Knowledge verification from data,'' \emph{IEEE Transactions on Neural Networks and Learning Systems}, pp. 1--15, 2022.

\bibitem{wang2023distribution}
X.~Wang, T.~Ban, L.~Chen, M.~Usman, T.~Wu, Q.~Chen, and H.~Chen, ``A distribution-based representation of knowledge quality,'' \emph{Knowledge-Based Systems}, vol. 281, p. 111054, 2023.

\end{thebibliography}

\vspace{-1cm}
\begin{IEEEbiography}[{\includegraphics[width=1in,height=1.25in,clip,keepaspectratio]{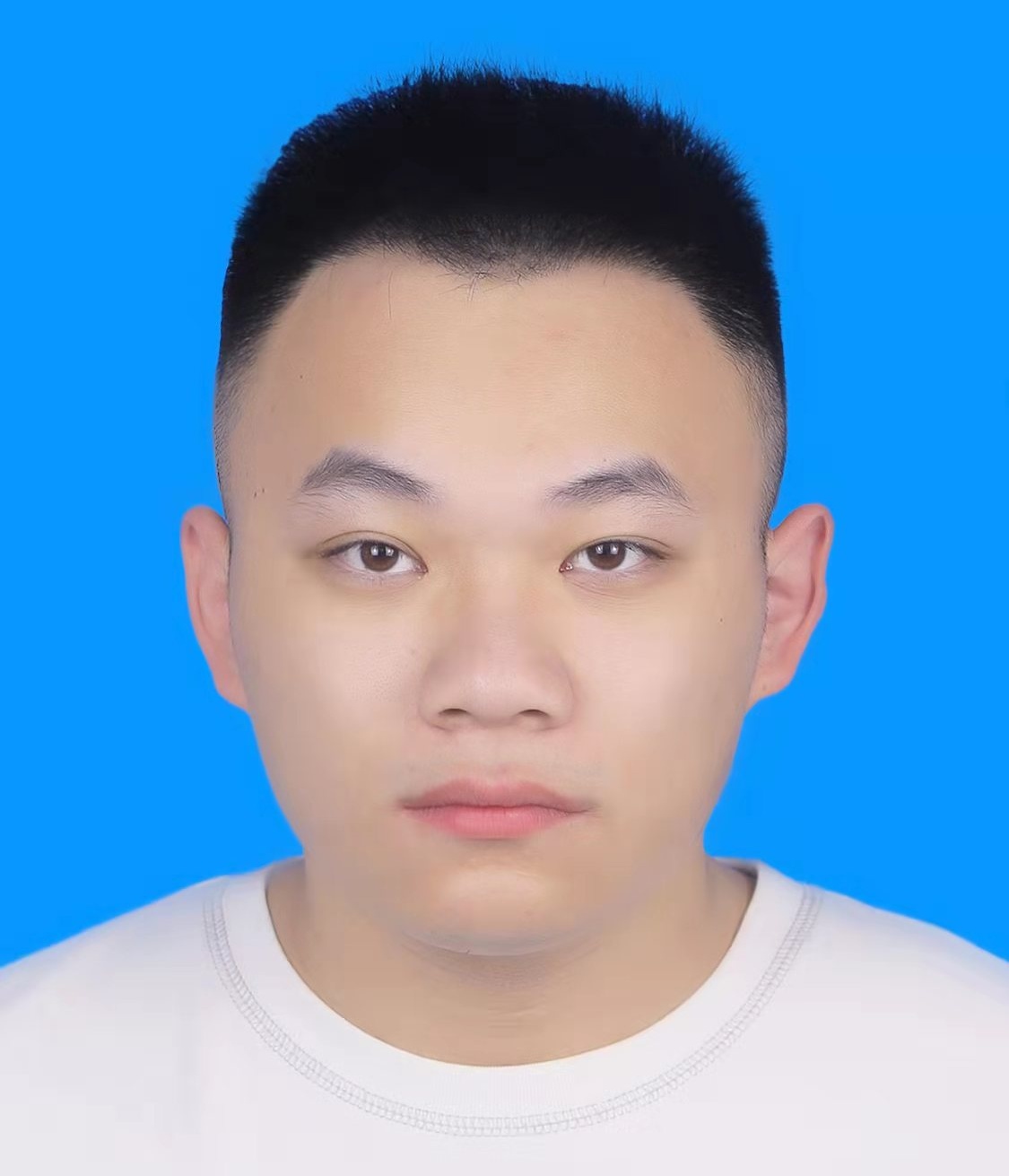}}]
  {Lyuzhou Chen} received the B.Sc. degree from the University of Science and Technology of China, Hefei, China, in 2020. He is currently pursuing the Ph.D. degree with the School of Data Science, University of Science and Technology of China, Hefei, China.

  His current research interests include causal learning, knowledge engineering and ensemble learning.
\end{IEEEbiography}
\vspace{-1cm}

\begin{IEEEbiography}[{\includegraphics[width=1in,height=1.25in,clip,keepaspectratio]{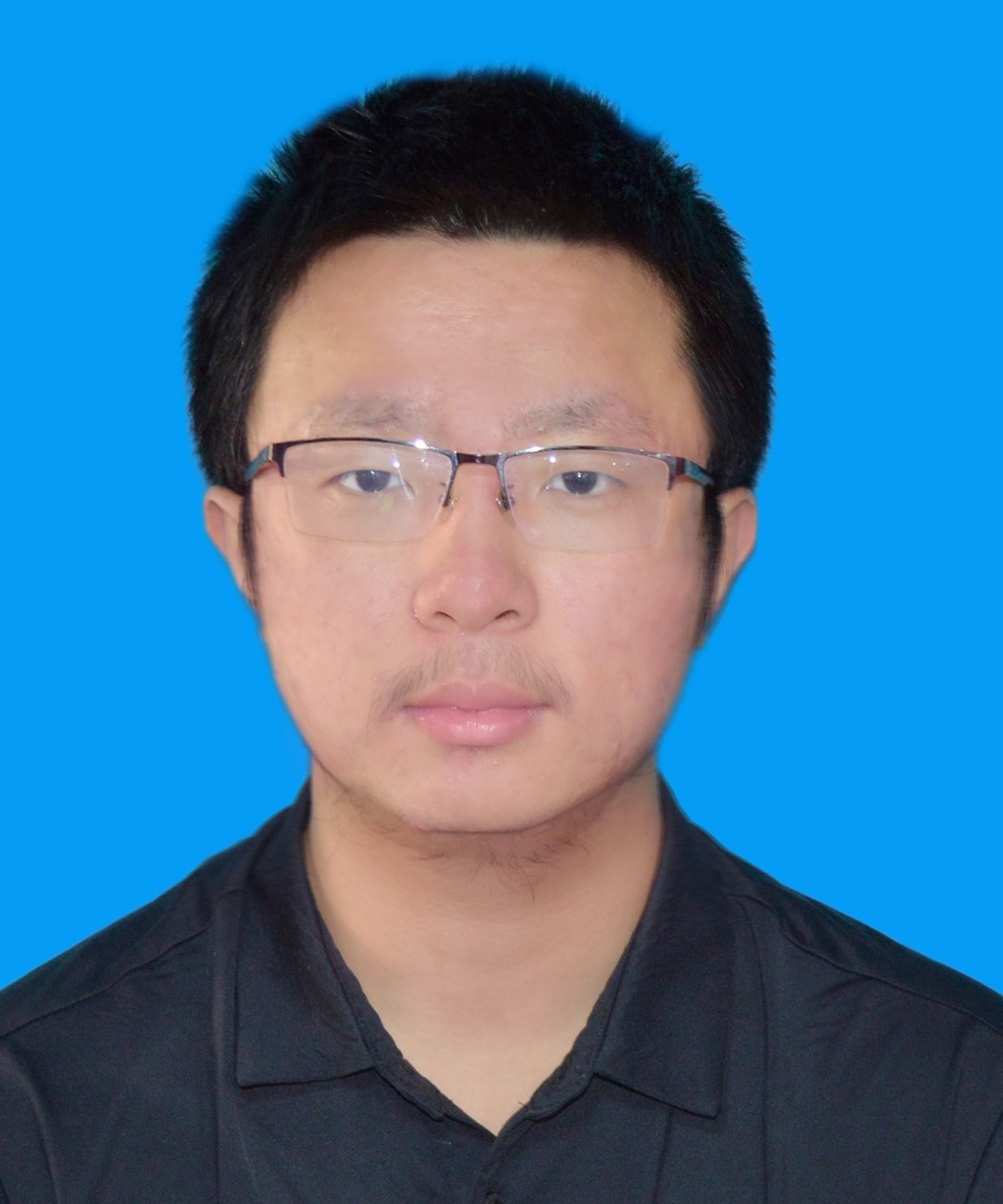}}]
  {Taiyu Ban} received the B.Sc. degree in Computer Science and Technology from the School of the Gifted Young, University of Science and Technology of China, Hefei, China, in 2020. He is currently pursuing the Ph.D. degree with the School of Computer Science and Technology, University of Science and Technology of China, Hefei, China.

  His current research interests include machine learning and knowledge engineering.
\end{IEEEbiography}
\vspace{-1cm}

\begin{IEEEbiography}[{\includegraphics[width=1in,height=1.25in,clip,keepaspectratio]{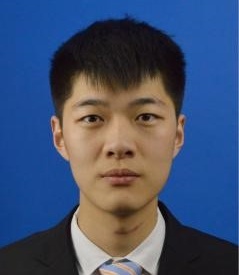}}]
  {Xiangyu Wang} received the B.Sc. degree from Donghua University, Shanghai, China, in 2015. He is currently pursuing the Ph.D. degree with the School of Data Science, University of Science and Technology of China, Hefei, China.

  His research interests include knowledge engineering and machine learning.
\end{IEEEbiography}
\vspace{-1cm}

\begin{IEEEbiography}[{\includegraphics[width=1in,height=1.25in,clip,keepaspectratio]{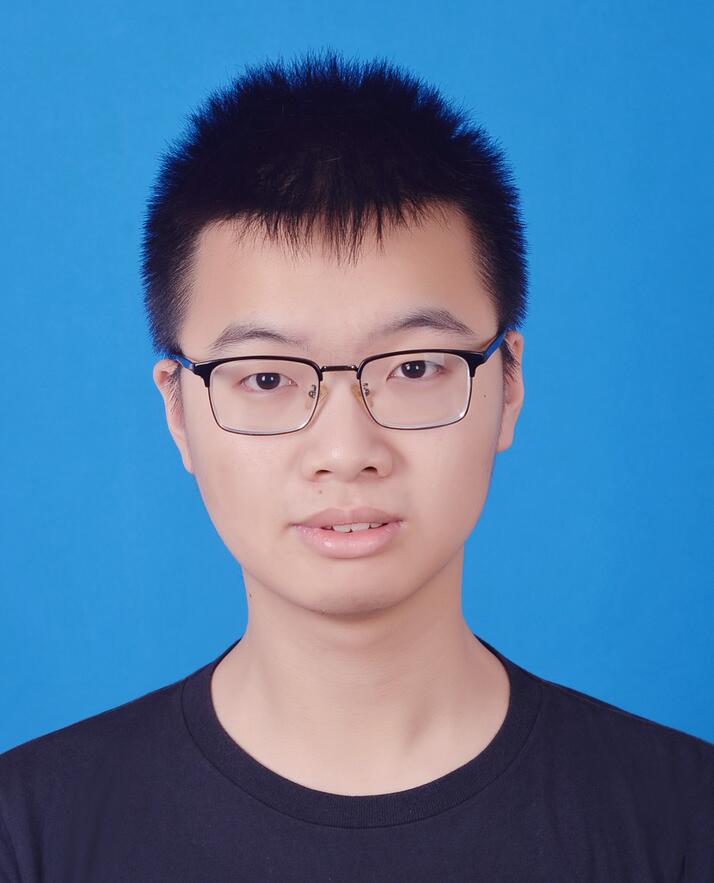}}]
  {Derui Lyu} received the B.Sc. degree in intelligence
  science and technology from the School of Artificial Intelligence, Xidian University, Xi’an, China, in 2021. He is currently pursuing the M.Sc. degree with the School of Computer Science and Technology, University of Science and Technology of China, Hefei, China.

  His current research interests include machine learning and knowledge engineering.
\end{IEEEbiography}
\vspace{-1cm}

\begin{IEEEbiography}[{\includegraphics[width=1in,height=1.25in,clip,keepaspectratio]{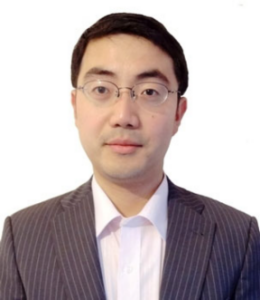}}]
  {Huanhuan Chen} (Senior Member, IEEE) received the B.Sc. degree from the University of Science and Technology of China (USTC), Hefei, China, and the Ph.D. degree in computer science from the University of Birmingham, Birmingham, U.K.
  He is currently a Full Professor with the School of Computer Science and Technology, USTC. His current research interests include neural networks, Bayesian inference, and evolutionary computation.

  He is now associate editor of the IEEE TRANSACTIONS ON NEURAL NETWORKS AND LEARNING SYSTEM, and the IEEE TRANSITION ON EMERGING TOPICS IN COMPUTATIONAL INTELLIGENCE.
\end{IEEEbiography}

\end{document}